\documentclass[ais]{iosart2x}

\usepackage{bm}
\usepackage{subcaption}
\usepackage[english, ruled, linesnumbered, vlined]{algorithm2e}
\usepackage{booktabs}

\usepackage{tabto}

\pubyear{0000}
\volume{0}
\firstpage{1}
\lastpage{1}

\begin{document}

\begin{frontmatter}

\title{Interactive Restriction of a Mobile Robot's Workspace in a Smart Home Environment}
\runningtitle{Interactive Restriction of a Mobile Robot's Workspace in a Smart Home Environment}
\subtitle{The final publication is available at IOS Press through \url{http://dx.doi.org/10.3233/AIS-190539}}

\author[A,B]{\inits{N.}\fnms{Dennis} \snm{Sprute}\ead[label=e1]{dennis.sprute@fh-bielefeld.de}%
\thanks{Corresponding author. \printead{e1}.}},
\author[B]{\inits{N.N.}\fnms{Klaus} \snm{T\"onnies}\ead[label=e2]{klaus@isg.cs.uni-magdeburg.de}}
and
\author[A]{\inits{N.-N.}\fnms{Matthias} \snm{K\"onig}\ead[label=e3]{matthias.koenig@fh-bielefeld.de}}
\runningauthor{D. Sprute et al.}
\address[A]{Campus Minden, \orgname{Bielefeld University of Applied Sciences}, Minden, \cny{Germany}\printead[presep={\\}]{e1,e3}}
\address[B]{Faculty of Computer Science, \orgname{Otto-von-Guericke University Magdeburg}, Magdeburg, \cny{Germany}\printead[presep={\\}]{e2}}


\begin{abstract}
Virtual borders are employed to allow humans the interactive and flexible restriction of their mobile robots' workspaces in human-centered environments, e.g. to exclude privacy zones from the workspace or to indicate certain areas for working. They have been successfully specified in interaction processes using methods from human-robot interaction. However, these methods often lack an expressive feedback system, are restricted to robot's on-board interaction capabilities and require a direct line of sight between human and robot. This negatively affects the user experience and interaction time. Therefore, we investigate the effect of a smart environment on the teaching of virtual borders with the objective to enhance the perceptual and interaction capabilities of a robot. For this purpose, we propose a novel interaction method based on a laser pointer, that leverages a smart home environment in the interaction process. This interaction method comprises an architecture for a smart home environment designed to support the interaction process, the cooperation of human, robot and smart environment in the interaction process, a cooperative perception including stationary and mobile cameras to perceive laser spots and an algorithm to extract virtual borders from multiple camera observations. The results of an experimental evaluation support our hypotheses that our novel interaction method features a significantly shorter interaction time and a better user experience compared to an approach without support of a smart environment. Moreover, the interaction method does not negatively affect other user requirements concerning completeness and accuracy.
\end{abstract}

\begin{keyword}
\kwd{Robot Workspace Restriction}
\kwd{Virtual Borders}
\kwd{Smart Home Environment}
\kwd{Network Robot System}
\end{keyword}

\end{frontmatter}

\section{Introduction}
\label{sec:intro}
Nowadays, physical environments become more and more smart comprising different kinds of sensors and actuators for the perception and manipulation of the environment. These devices are connected via a network with each other, and in combination with intelligent software, such a smart environment is able to provide context-aware services to humans resulting in an intelligent environment~\cite{Augusto:2013}. A certain form of a smart environment, which we deal with in this work, is a smart home equipped with computing and information technology providing services to residents~\cite{Aldrich:2003}. When additionally integrating a robot into a smart environment, this combination is referred to a network robot system (NRS), which extends the perceptual and interaction capabilities of the robot and environment~\cite{Sanfeliu:2008}. This allows especially mobile robots, which are able to move in the environment using their locomotion system~\cite{Corke:2011}, to provide complex services to the residents, e.g. providing sophisticated healthcare services~\cite{Li:2013}, tidying up rooms~\cite{Rasch:2019} or supporting humans in a kitchen~\cite{Rusu:2008}.  The locations, that can be reached using the locomotion systems, are defined as the mobile robot's workspace.

Although residents appreciate the services of mobile robots in their home environments, there are scenarios in which humans want to restrict the workspaces of their mobile robots, i.e. they want to define restriction areas. For example, restriction areas are needed (1)~to exclude intimate rooms, such as bed- or bathrooms, due to privacy concerns~\cite{Ziefle:2011}, (2)~to exclude carpet areas from the mobile robot's workspace to avoid navigation errors~\cite{Hawes:2017} or (3)~to indicate certain areas for working, such as spot vacuum cleaning~\cite{Gromov:2019}. These scenarios can be summarized to the problem, that we deal with in this work: \textit{the restriction of a 3-DoF mobile robot's workspace and change of its navigational behavior according to the humans' needs}. A 3-DoF mobile robot operates on the ground plane and has three degrees of freedom, i.e. 2D position and orientation. The described problem is highly relevant for users living in human-robot shared spaces, such as home environments. Since a restriction area cannot be directly perceived by the mobile robot due to computational or perceptual limitations and/or explicit knowledge of a human is required, an interaction process between human and robot is necessary. This interaction process has to (1)~allow a transfer of spatial information concerning the restriction areas and (2)~has to provide a feedback channel to inform the human about the progress of the interaction process and its results.

Current state-of-the-art solutions focus on methods from the field of human-robot interaction (HRI) to specify restriction areas. For this purpose, the concept of a virtual border is employed, which is a non-physical border not directly visible to the user but that effectively and flexibly restricts a mobile robot's workspace~\cite{Sprute:2017a}. Thus, a mobile robot changes its navigational behavior and respects the user-defined restriction areas. In order to specify a virtual border in an interaction process, a user employs an interaction method, that is built around a user interface for interaction. Current interaction methods are either based on visual displays or mediator-based pointing gestures, e.g. using a laser pointer as pointing device~\cite{Sprute:2018c}. Especially, the latter one is a natural and intuitive method of non-verbal communication making it applicable by non-expert users~\cite{Sprute:2019a}. These non-expert users are residents of home environments and do not have much experience with robots but can interact with common consumer devices, such as tablets or smartphones.

However, despite of the natural and intuitive interaction using pointing gestures, this category of interaction methods requires a direct line of sight between human and robot's on-board camera. Due to the limited field of view of mobile robots' cameras, this yields a limited interaction space affecting the quality of interaction negatively, e.g. in terms of interaction time. Moreover, the feedback capabilities are limited to the mobile robot's on-board components, e.g. non-speech audio~\cite{Kim:2009} or colored lights giving feedback~\cite{Baraka:2018}. This also affects the quality of interaction negatively, e.g. in terms of user experience. Therefore, in this work our objective is the investigation of the role of a smart home environment in the interaction process to overcome these limitations. We hypothesize that the additional components of a smart home environment can improve the interaction time and user experience compared to a solution without support of a smart environment because additional sensors and actuators increase the perception and interaction capabilities. Moreover, we hypothesize that other user requirements, such as accuracy and completeness of the interaction method~\cite{Sprute:2018c}, will not be negatively affected.

Based on this objective, we contribute the following aspects to the state of the art. We investigate the effect of a smart environment on the interaction process of restricting a mobile robot's workspace. Therefore, we propose a novel interaction method incorporating a smart home environment and laser pointer as interaction device. This interaction method encompasses (1)~an architecture for a smart environment designed to support the interaction process, (2)~the cooperation of human, robot and smart environment in the interaction process, (3)~a cooperative perception including stationary and mobile cameras to perceive laser spots and (4)~an algorithm to extract virtual borders from multiple camera observations. 

The remainder of this paper is structured as follows: in the next section, we give an overview of related works concerning the restriction of mobile robots' workspaces, the integration of robots into smart environments and interaction opportunities of network robot systems. Based on our objective and contributions of related works, we point out a research gap and formulate three open research questions as basis for the remainder of this work. Afterwards, we first formally define a virtual border as a concept to flexibly define restriction areas. This is the basis for our novel interaction method addressing the problem of restricting a mobile robot's workspace. Details on the interaction method and how it addresses the research questions are given in Sect.~\ref{sec:approach}. Subsequently, we evaluate the interaction method in comparison to a baseline method with the focus on the test of our hypotheses. Finally, we conclude our work, point out current limitations and suggest future work.

\section{Related Work}
\label{sec:relatedWork}
\subsection{Mobile Robot Workspace Restriction}
As already stated in the introduction, current solutions to restrict the workspace of a mobile robot are based on sole HRI without support of a smart environment. Commercial solutions comprise magnetic stripes placed on the ground~\cite{Neato:2019} and virtual wall systems based on beacon devices~\cite{Chiu:2011}, that emit infrared light beams. Despite their effectiveness, they are intrusive, i.e. additional physical components on the ground are necessary to restrict the workspace, and inflexible, i.e. restriction areas are limited to certain sizes and shapes. Thus, they are not applicable for our problem because the considered scenarios require restriction areas of arbitrary sizes and shapes. An alternative is sketching restriction areas on a map of the environment containing occupancy probabilities~\cite{Ackerman:2017}. These maps are created by today's home robots and are used for navigation purposes. However, this interaction method is inaccurate, i.e. there are strong inaccuracies between the user-defined restriction areas as a result of an interaction process and the restriction areas to be intended for restriction by a human. This is caused by a correspondence problem between points on the map and in the environment~\cite{Sprute:2018c}. In order to address this shortcoming, Sprute et al. introduced a framework for interactive teaching of virtual borders based on robot guidance~\cite{Sprute:2017a}. This allows a user to guide a mobile robot, e.g. using a laser pointer as interaction device, and specify a virtual border as the robot's trajectory~\cite{Sprute:2019a}. Although this interaction is accurate due to the accurate localization of the mobile robot, the framework suffers from a linear interaction time regarding the border length and lacks an expressive feedback system. This is caused by the requirement concerning a direct line of sight between human and robot and limited on-board feedback capabilities. A comprehensive user study dealing with different HRI methods for teaching virtual borders revealed augmented reality as the most powerful interface for the given task~\cite{Sprute:2018c}. However, this approach requires specialized hardware, i.e. a RGB-D tablet, which limits the potential number of users. Nonetheless, the user study showed that 36\% of the participants, i.e. the second largest group, also preferred a laser pointer as interaction device due to its simplicity, comfort and intuitiveness during interaction. These advantages of a laser pointer are also revealed in other robot tasks, such as guiding a mobile robot to a 3D location~\cite{Kemp:2008}, controlling a robot using stroke gestures~\cite{Ishii:2009} and teaching visual objects ~\cite{Rouanet:2013}.

\subsection{Network Robot Systems}
In order to benefit from the advantages of laser pointers and reduce their limitations caused by the requirement concerning a direct line of sight for interaction and limited on-board feedback capabilities, we investigate the incorporation of a smart home environment with additional sensors and actuators into the interaction process. This integration of robots into smart environments is known as network robot systems (NRS)~\cite{Sanfeliu:2008}. Such a system is characterized by five elements: physical embodiment, autonomous capabilities, network-based cooperation, environment sensors and actuators and human-robot interaction. Other related terms are ubiquitous robots~\cite{kim:2007}, physically embedded intelligent systems (PEIS)~\cite{saffiotti:2008}, Kukanchi~\cite{nor:2014} and informationally structured environment~\cite{pyo:2015}. This ubiquitous robotics paradigm is leveraged to provide more complex and more efficient robot services to humans. For example, Sandygulova et al. developed a portable ubiquitous robotics testbed consisting of a kitchen equipped with a wireless sensor network, a camera and mobile robot~\cite{Sandygulova:2016} and Djaid et al. integrate a robot wheelchair with a manipulatable arm into an intelligent environment~\cite{Djaid:2017}. Other applications comprise the improvement of a mobile robot's object search~\cite{Sprute:2017b}, the development of a human-aware task planner based on observations of the smart environment~\cite{Cirillo:2010} and a gesture-based object localization approach for robot applications in intelligent environments~\cite{Sprute:2018a}. All these approaches show that the cooperation between robots and a smart environment can improve the quality of robot applications, e.g. through enhanced perception and interaction abilities~\cite{Simoens:2018}. 

\subsection{Interaction in Smart Environments}
However, only a single attempt was made to employ a smart environment to restrict the workspace of a mobile robot, i.e. Liu et al. use several bird's eye view cameras mounted in the environment to visualize a top view of the environment on a tablet's screen on which the user can draw arbitrary restriction areas~\cite{Liu:2011}. But this approach relies on a full camera view coverage of the environment and does not deal with partial occlusions, e.g. by furniture. Therefore, it is not applicable to our problem because a home environment is typically not fully covered by cameras due to privacy concerns~\cite{Apthorpe:2018}. Nonetheless, there are already works dealing with the recognition of laser pointer gestures using cameras of a smart environment, e.g. Shibata et al. propose a laser pointer approach to allow humans to control a mobile robot~\cite{Shibata:2011}. Similarly, Ishii et al. use cameras integrated into the environment for laser spot tracking allowing a human to define stroke gestures with a laser pointer to control a network-connected mobile robot~\cite{Ishii:2009}. These approaches are no more limited to the mobile robot's field of view and thus increase the interaction space. However, the gestures are exclusively recognized by cameras integrated in the smart environment. Hence, the interaction strongly depends on the number of cameras in the environment and their fields of view. There is no work considering a cooperative behavior employing cameras from the smart environment \textit{and} the mobile robot for this task. This would be an opportunity to overcome issues concerning camera view coverage and occlusions. Moreover, smart environments provide additional visualization capabilities that could be used to provide feedback to the human, e.g. visual displays integrated into the environment~\cite{Butz:2010}\cite{Kang:2018}. 

\subsection{Research Gap}
These findings lead us to the general question, we want to answer in this work: \textit{how can a smart home environment improve the interaction process of restricting a mobile robot's workspace using a laser pointer with respect to the interaction time and user experience?} This question involves research questions of (1)~which sensors and actuators of a smart environment can be used to benefit the interaction process, (2)~how to realize a cooperation of human, robot and smart environment in the interaction process and (3)~how to cooperatively perceive and combine multiple camera observations of laser points to restrict the mobile robot's workspace. These questions are the basis for the remainder of this work, which will be answered by prototypically implementing a solution and an empirical evaluation.

\section{Virtual Borders}
\label{sec:virtualBorders}
Before we present our solution to the given problem, we first introduce the concept of a virtual border in more detail. As already mentioned, this is a non-physical border not directly visible to a human but respected by a mobile robot during navigation. It comprises spatial information necessary to define a restriction area in an interaction process. Thus, a virtual border can be used to interactively specify restriction areas, such as carpets or privacy zones. This concept was already formally defined as a triple $V = (\mathcal{P}, \bm{s}, \delta)$ in a previous work~\cite{Sprute:2018b}. The three components of a virtual border are described below:
\begin{enumerate}
	\item \textbf{Virtual border points} $\mathcal{P}$: These are $n > 1$ points $\bm{p_i} \in \mathbb{R}^2$ with $1 < i < n$ organized as a polygonal chain, that specifies a boundary of a restriction area on the ground plane. There are two types of virtual borders: (1)~a closed polygonal chain (\textit{polygon}) and (2)~a simple polygonal chain (\textit{separating curve}). The former one divides an environment into an inner and outer area, while the latter one does not directly partition an environment. Thus, its first and last line segments are linearly extended to the physical borders of the environment to allow the separation of the area. This formulation allows the definition of restriction areas with arbitrary sizes and shapes.
    \item \textbf{Seed point} $\bm{s}$: This is a point on the ground plane $\bm{s} \in \mathbb{R}^2$ indicating the area to be modified by the human in the interaction process, i.e. the area to be restricted.
    \item \textbf{Occupancy probability} $\delta$: This component specifies the occupancy probability of the area to be modified as indicated by the seed point $\bm{s}$.\footnote{In this work, we only consider 100\% to indicate occupied areas, but in the future this component could be used to model levels of restriction. For example, an occupancy of 75\% could mean that a mobile robot should avoid this area unless it is necessary.} 
\end{enumerate}
In order to enforce a mobile robot to change its navigational behavior, a virtual border has to be integrated into a map of the environment. Such a map is used by mobile robots to autonomously navigate in the environment, i.e. calculating a path from a robot's start to a goal pose considering physical obstacles~\cite{Minguez:2016}. For this purpose, a 2D occupancy grid map (OGM)~\cite{Moravec:1985} representation is chosen due to its popularity in robot navigation and guidance. It models the physical environment, i.e. walls or furniture, in terms of discrete cells containing occupancy probabilities, i.e. \textit{free}, \textit{occupied} and \textit{unknown} areas. To integrate a virtual border into a map of the environment, we employ a map integration algorithm~\cite{Sprute:2018b}. The input is a map of the physical environment $M_{prior}$ and a user-defined virtual border $V$. The output of the map integration algorithm is a 2D OGM $M_{posterior}$ containing physical as well as the user-defined virtual borders. Thus, when considering $M_{posterior}$ as basis for navigation, a mobile robot changes its navigational behavior according to the user-defined restrictions. Furthermore, the algorithm can be iteratively applied with different virtual borders $V^{*}=\{V_1, V_2, ..., V_N\}$, i.e. the output of the \mbox{$i$-th} interaction process becomes the input of the \mbox{$i+1$-th} interaction process. This map integration algorithm has already been proven to be correct, i.e. it changes the mobile robot's navigational behavior, and flexible, i.e. a human can specify arbitrary virtual borders~\cite{Sprute:2018c}. For this reason, we also employ the algorithm in this work to allow users the definition of restriction areas.

\section{Workspace Restriction in a Smart Home}
\label{sec:approach}
Due to the advantages of mobile robot applications integrated into smart environments and the drawbacks of current solutions for our problem, i.e. the restriction of a mobile robot's workspace and change of its navigational behavior, we take up the opportunity of an enhanced perception and interaction capability provided by a smart home environment in this work. To this end, we first present a smart home design and explain how smart home components can be leveraged to address limitations of current solutions. Subsequently, we describe how these smart home components are incorporated into the interaction process between human and robot. Thus, this cooperative behavior results in a novel interaction method based on human-robot-environment interaction. Finally, we present details on the cooperative perception based on smart home's and mobile robot's cameras, which is a fundamental part of the proposed interaction method. It is the goal to combine multiple camera observations of laser points and extract a single virtual border from these observations. For this purpose, we developed a novel multi-stage algorithm addressing several challenges.

\subsection{Smart Environment Design}
\label{sec:desgin}
As stated in Sect.~\ref{sec:relatedWork}, laser pointers are quite popular for human-robot interactions involving the transfer of spatial information due to their natural interaction mimicking human pointing gestures. For example, our baseline method for the restriction of a mobile robot's workspace employs a laser pointer as interaction device~\cite{Sprute:2019a}. Nonetheless, this kind of interaction features some drawbacks. The major drawback of the existing interaction method is the requirement concerning a direct line of sight between human and robot when specifying virtual borders. Thus, the mobile robot has to follow the laser spot, but it is restricted by velocity constraints, which lead to a relatively long interaction time. Moreover, only limited feedback about the current state of the interaction process can be conveyed using simple on-board LEDs and non-speech audio sound, e.g. the robot provides a beep sound whenever a laser spot is detected. However, no complex feedback concerning the spatial information of the specified virtual borders can be provided to the human using these communication channels. Another drawback of the baseline approach is the interaction to change between different states of the interaction process, e.g. specifying virtual border points or the seed point. For this purpose, visual codes generated by the laser pointer or push buttons on the mobile robot are provided. But visual codes can be error-prone due to changing light conditions and interaction using buttons requires the user to be in the vicinity of the robot. 

\begin{figure}
	\centering
	\includegraphics[width=0.98\columnwidth]{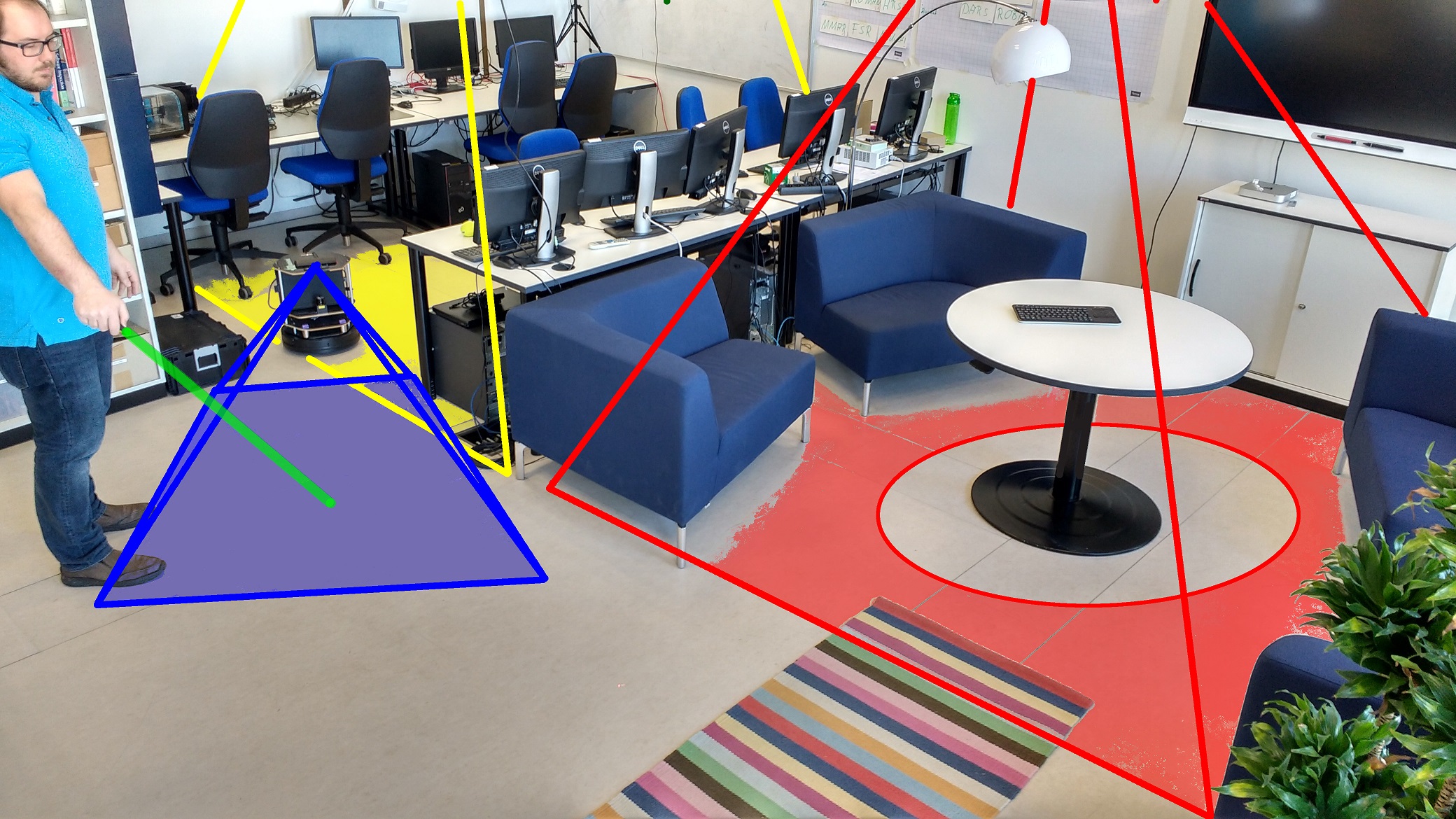}   
	\caption{A person defines a virtual border in the environment using a laser pointer. The spot is observed by stationary cameras in the environment (yellow and red) and a mobile camera on a robot (blue). A smart display (top right) provides visual feedback of the complex spatial information, and a smart speaker facilitates interaction.}
	\label{fig:motivation}
\end{figure}
Addressing these shortcomings of the baseline approach, we propose the incorporation of additional smart home components in the interaction process as the answer to our first research question. This deals with the question of which sensors and actuators of the smart environment can be used to benefit the interaction process. Our proposed smart home environment is shown in Fig.~\ref{fig:motivation} and mainly consists of three components: (1)~a camera network comprising stationary RGB cameras, (2)~a smart display and (3)~a smart speaker allowing the processing of voice commands. The first component, i.e. the camera network, is intended to increase the perceptual capabilities in combination with the mobile robot's on-board camera. Stationary cameras are integrated (yellow and red fields of view) to cover certain areas of the environment, while a mobile camera mounted on a robot (blue field of view) can observe areas that are not covered by the stationary cameras due to their installation or occlusions, e.g. under the table. Hence, this combination allows perception even if the stationary cameras do not cover all areas of the environment, which is typically the case in smart home environments. The second component, i.e. the smart display, is intended to provide expressive feedback to the user by visualizing the progress and result of the interaction process. Finally, it is the idea to employ a smart speaker to facilitate the change of different states of the interaction process using voice commands.

\subsection{Human-Robot-Environment Interaction}
In order to achieve our objective of a reduced interaction time and increased user experience compared to the baseline approach, we propose a new interaction method leveraging the combination of a mobile robot and the described smart home environment in the interaction process. We denote this combination as network robot system (NRS) in the following. This interaction method is intended to answer our second research question of how to realize a cooperation of human, robot and smart environment in the interaction process. As pointed out in the introduction, the interaction process between human and robot (1)~has to allow a transfer of spatial information from human to robot and (2)~has to provide feedback about the interaction process from robot to human. The first property is addressed by allowing a user to specify virtual borders by ``drawing'' directly in the environment using a common laser pointer. A laser spot is cooperatively perceived by the stationary camera network and the mobile robot's on-board camera. 

The interaction method comprises several internal states, that reflect the three components of a virtual border as described in Sect.~\ref{sec:virtualBorders}. The states are described as follows:
\begin{itemize}
	\item \textbf{Default}: The NRS is in an inactive state and ignores all user interactions.
    \item \textbf{Border}: The NRS recognizes laser spots that are used to specify virtual border points~$\mathcal{P}$. If the stationary cameras perceive a user's laser spot, the system automatically sends a mobile robot to this area. Thus, the mobile robot autonomously navigates to this area and can act as mobile camera if the stationary cameras lose track of the laser spot.
    \item \textbf{Seed}: The NRS recognizes laser spots and calculates the seed point $\bm{s}$. Similar to the \textit{Border} state, a mobile robot simultaneously moves to the laser spot position if a stationary camera perceives a laser spot. The seed point $\bm{s}$ indicates the restriction area, i.e. $\delta=1$.
    \item \textbf{Guide}: This state is similar to the \textit{Default} state, but the user can guide the mobile robot using the laser pointer without storing the laser spots.
    The state should be never reached if at least one of the stationary cameras' fields of view covers a part of the restriction area and can send the mobile robot to this area. However, we incorporate this state into our interaction method to ensure its functionality in case of an absence of stationary cameras. In this case, a user has to manually guide the mobile robot to the restriction area because the robot cannot be automatically sent to the area due to missing information of the smart environment. Hence, the system degenerates to the baseline approach, i.e. a system without support of a smart home environment.
\end{itemize}
In order to switch between the different states, a user can employ the following speech commands, that are perceived and processed using the smart speaker integrated into the smart home:
\begin{itemize}
	\item \textbf{Define border}: This command is used to start the specification of virtual border points $\mathcal{P}$, thus switching to state \textit{Border}.
    \item \textbf{Define seed}: This command is used to start the specification of a seed point $\bm{s}$, thus switching to state \textit{Seed}.
    \item \textbf{Guide robot}: The system's internal state switches to \textit{Guide} so that a user can guide the mobile robot using the laser pointer.
    \item \textbf{Save}: This command is employed when a user wants to integrate and save his or her user-defined virtual border into the map of the environment.
    \item \textbf{Cancel}: If a user does not want to save his or her user-defined virtual border, he or she can cancel the interaction process. Hence, the internal state changes to \textit{Default}.
\end{itemize}
Finally, we realize the second property of an interaction process, i.e. a feedback channel from robot to human, by extending the baseline's feedback system. In addition to mobile robot's non-speech audio sound and colored light feedback indicating state changes and the detection of a laser spot, we use the smart display integrated into the environment to provide more complex feedback. This includes the visualization of the 2D OGM of the environment, the mobile robot's current position on the map and the progress of the spatial information transfer. After successfully accomplishing an interaction process, the mobile robot's workspace containing the user-defined virtual borders is also shown on the display. 

\subsection{Cooperative Perception}
While the incorporation of the smart display and smart speaker in the interaction method is straightforward, the incorporation of the camera network, which is used to increase the interaction space, is more challenging. There are mainly two reasons: (1)~multiple cameras, stationary and mobile, have to be integrated into an architecture that supports the interaction process and (2)~a single virtual border has to be extracted from multiple camera observations including noisy data.

\subsubsection{Architecture}
\begin{figure}
	\centering
	\includegraphics[width=\columnwidth]{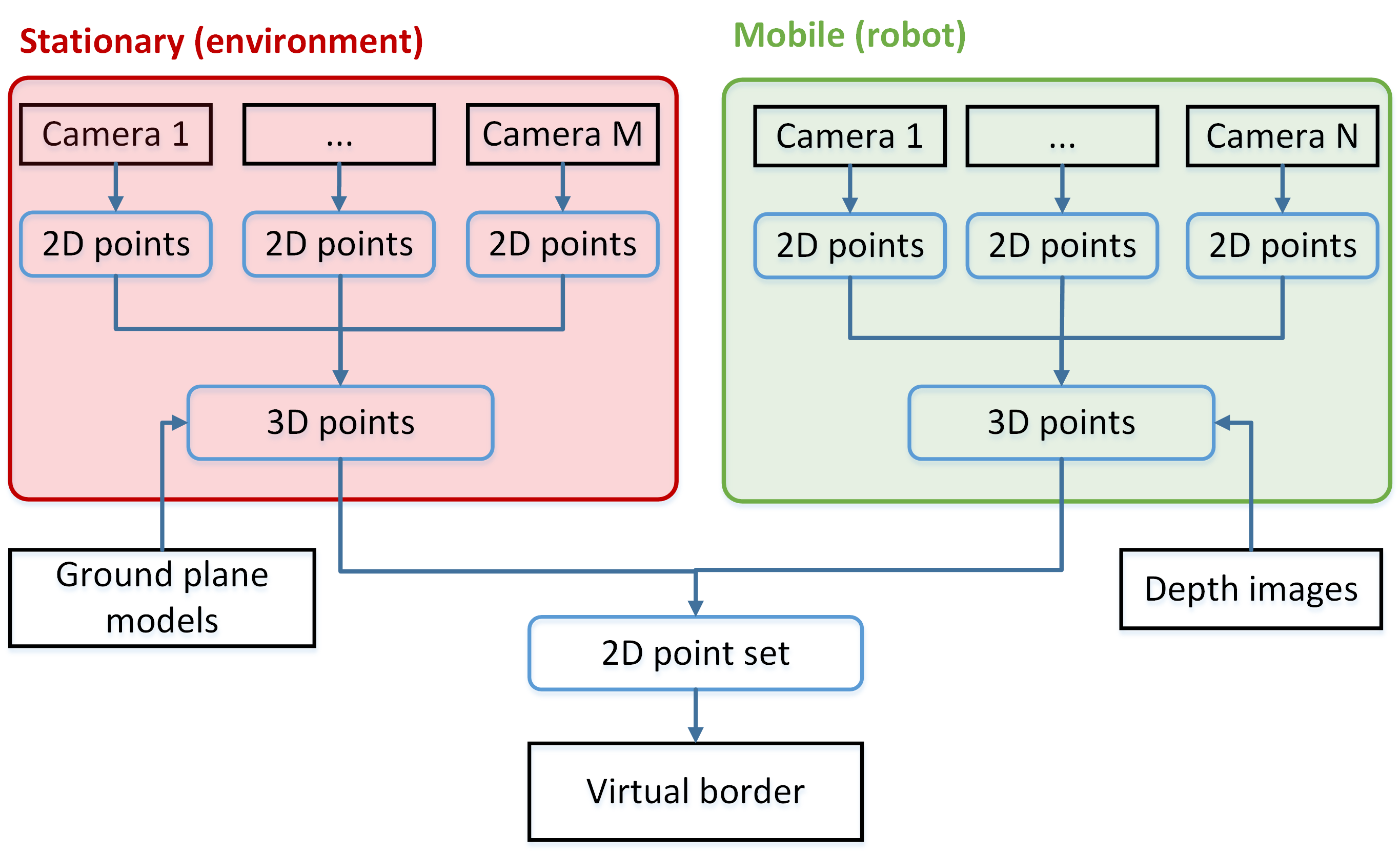}   
	\caption{Architecture of the cooperative perception for specifying virtual borders based on multiple camera views.}
	\label{fig:overview}
\end{figure}
Addressing the first aspect, we propose the architecture, that is illustrated in Fig.~\ref{fig:overview}, as a first part of our answer to the third research question of how to cooperatively perceive and combine multiple camera observations to restrict the mobile robot's workspace. The architecture consists of $M$ stationary cameras integrated in the environment and $N$ mobile cameras on mobile robots\footnote{Although we consider a single mobile robot in this work, we designed the architecture with multiple mobile cameras due to scalability options in the future.}. Each camera independently performs laser point detection in image space resulting in a 2D point $\bm{p} \in \mathbb{R}^2$ for each detected laser spot. We apply the laser point detection algorithm from the baseline approach that is based on illumination and morphologic properties of a laser spot, i.e. circular shape, specific size and extreme brightness compared to its local environment~\citep{Sprute:2019a}. Subsequently, each point $\bm{p}$ is projected into 3D world space $\bm{P} \in \mathbb{R}^3$ using either a ground plane model in case of the stationary cameras or an additional depth image in case of the mobile cameras. The aim is to make laser point observations independent of the cameras. Since the resulting points are points on the ground plane, they degenerate to 2D positions with respect to the map coordinate frame. This coordinate frame is the origin of the environment's map and does not contain a third dimension because we focus on 3-DoF mobile robots. All camera transformations with respect to the map coordinate frame are known to ensure transformations from the cameras' to the map's coordinate frame. These transformations belong to the Special Euclidean group $SE(3)$ and are determined in advance during installation. In case of a mobile robot, this transformation is dynamically adapted according to the localization of the robot in the environment. After projecting the points into the map coordinate frame, all points are incorporated into a single 2D point set describing all laser point detections in space independent of the camera source. Finally, a virtual border is extracted from this point set.

\subsubsection{Virtual Border Extraction}
\begin{figure*}
	\centering
	\includegraphics[width=\textwidth]{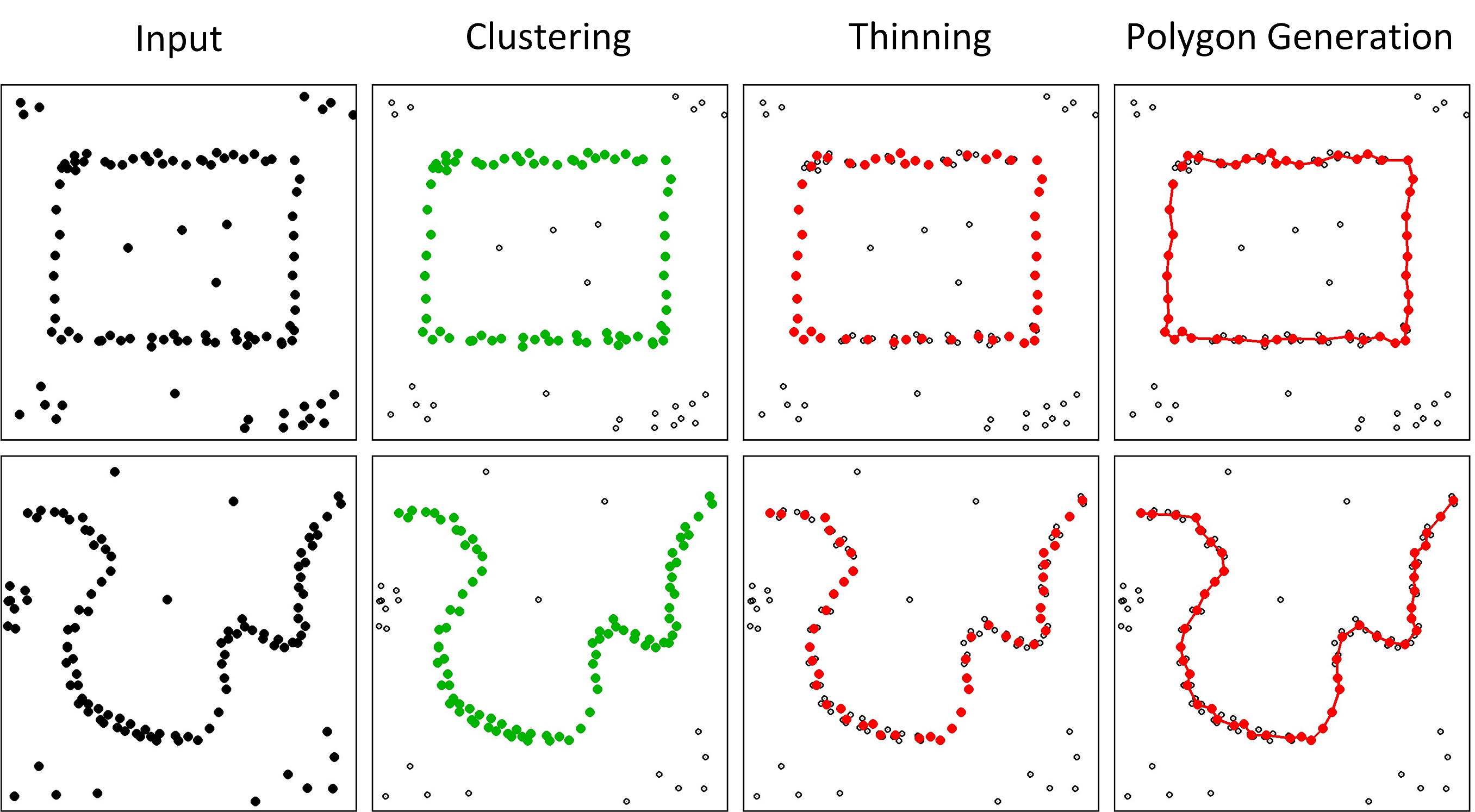}   
	\caption{Processing stages extracting a polygonal chain from a point set including noise. Each row corresponds to a user-defined point set, a polygon in the first and a separating curve in the second row. The first column visualizes the input point set containing virtual border points but also noise and other clusters. The points assigned to the virtual border cluster are colored green in the second column. Thinning the virtual border cluster yields the red point set in column three. This is used to generate a polygonal chain as shown in the last column.}
	\label{fig:polyGeneration}
\end{figure*}
While the extraction of the seed point~$\bm{s}$ is not challenging, the extraction of the polygonal chain~$\mathcal{P}$ includes several challenges that need to be adequately addressed. For example, the 2D point set acquired from multiple camera observations contains data points of a single user-defined polygonal chain~$\mathcal{P}$. However, this point set also contains (1)~noisy data points and possibly (2)~additional clusters. These can occur due to errors in the laser point detection algorithm or other areas in the environment, that have the same characteristics as a laser point. Moreover, the data points can be (3)~spatially redundant since the points are obtained from different cameras that may have an overlap of their fields of view, e.g. two overlapping static cameras of the environment or an overlap between a static camera and the mobile robot's camera. Moreover, calibration inaccuracies of the cameras and localization errors of the mobile robot can lead to (4)~inaccurate user-defined points. Finally, the (5)~generation of a polygonal chain from the point set is challenging because the polygonal chain can have an arbitrary shape. We address these challenges with a novel multi-stage virtual border extraction algorithm, which is the second part of our answer to the third research question. Fig.~\ref{fig:polyGeneration} illustrates the stages of the algorithm with two exemplary polygonal chains in the first and second row. We reference this figure throughout this subsection to explain the multi-stage algorithm.

\paragraph{Clustering}
The first stage of the extraction algorithm is the clustering stage as denoted in Algorithm~\ref{algo:clustering}. The input is a 2D point set $pointsetIn$ as shown in the first column of Fig.~\ref{fig:polyGeneration}, and the data points belonging to the polygonal chain $pointsetOut$ are the result. This stage is designed to addresses the first two challenges of the virtual border extraction step, i.e. extracting of the user-defined polygonal chain and discarding noisy data points and irrelevant clusters. A cluster is a group of spatially nearby data points, while noisy data points are single data points that do not belong to a cluster. Irrelevant clusters are characterized by certain expansion characteristics, that indicate the spatial expansion of the cluster measured as Euclidean distance between the diagonal points of the cluster's minimum bounding box. Due to the flexibility of a virtual border, i.e. a user can define arbitrary shapes, we applied the \mbox{DBSCAN}~\cite{Ester:1996} algorithm for clustering the data points (l.~\ref{line:dbscan}). This is a density-based clustering algorithm that can find clusters with different shapes and sizes. It is parameterized by $eps$ to define the distance threshold for a neighboring point and $minPts$ to define a core point. This is a point that has at least $minPts$ points within its distance $eps$. The result of the DBSCAN algorithm is a set of $clusters$ where each point of $pointsetIn$ is assigned to a cluster. Noisy data points, that do not belong to a cluster, are discarded. Afterwards, the algorithm selects the largest cluster with certain expansion characteristics defined by $minExp$ and $maxExp$. To this end, we order the clusters by their sizes (number of points) in descending order (l.~\ref{line:order}) to iterate over the clusters beginning with the largest cluster (l.~\ref{line:loopStart}ff.). The additional parameter $minSize$ is a lower threshold for the size of a cluster to exclude small clusters due to noise. In each iteration, it is checked if the expansion of the current cluster $c$ lies within the expansion thresholds $minExp$ and $maxExp$ to ignore irrelevant clusters (l.~\ref{line:exp}). The first cluster, that fulfills this condition, is returned as cluster of the polygonal chain. The result is visualized in the second column of Fig.~\ref{fig:polyGeneration} as green points. The black points are either noise or irrelevant clusters.
\begin{center}
	\begin{algorithm}       
        \LinesNumbered        
		\SetFuncSty{textbf}
        \SetKwInOut{KwParams}{Params}   
		\SetKwFunction{order}{orderClustersBySize}        
		\SetKwFunction{dbscan}{DBSCAN} 
		\SetKwFunction{expansion}{expansion}          	
  		\KwIn{pointsetIn}
        \KwOut{pointsetOut}
        \KwParams{eps, minPts, minExp, maxExp, minSize}
        \SetKwProg{func}{Function}{}{}  
        \SetKw{break}{break}
        \SetKw{in}{Input}
        \SetKw{out}{Output}
        \SetKw{params}{Params}
  		\func{clustering(\in, \out, \params)}
  		{
        	pointsetOut = $\emptyset$\;
  			clusters = \dbscan(pointsetIn, eps, minPts)\;\label{line:dbscan}
  			clusters = \order(clusters, minSize)\; \label{line:order}	
            \ForEach{c in clusters}
            {\label{line:loopStart} 
            
              \If{minExp < \expansion(c) < maxExp}
              {\label{line:exp}
              	pointsetOut = c\;
                \break\; 
              }  			 		
            }
  		}      
    \caption{Clustering step.}
    \label{algo:clustering}        
    \end{algorithm}
\end{center}

\paragraph{Thinning}
The second stage of the algorithm is the thinning stage (Algorithm~\ref{algo:thinning}), that reduces the number of points in the largest cluster from the previous stage. It is designed to remove spatially redundant data points and to smooth data points due to localization errors and calibration inaccuracies. Thus, it addresses the third and fourth challenge of the virtual border extraction step. For this purpose, the thinning algorithm identifies spatially nearby data points and replaces them by their mean value. To this end, the point $p$ with most neighbors within a distance $maxNeighborDist$ is selected~(l.~\ref{line:getPoint}) and its neighboring points $n$ are determined~(l.~\ref{line:neighbors}). If there is at least one neighboring point~(l.~\ref{line:neighborAvailable}), the mean point is calculated for these points $p \cup n$~(l.~\ref{line:mean}). Afterwards, these points are removed from the initial $pointsetIn$~(l.~\ref{line:remove}) and the mean point is added to $pointsetOut$~(l.~\ref{line:add}). In case that no data point has at least one neighboring point~(l.~\ref{line:neighborNotAvailable}), the iterative procedure terminates. Finally, all remaining points contained in $pointsetIn$, i.e. points without neighbors, are added to $pointsetOut$~(l.~\ref{line:union}), i.e. the set containing the thinned points. Thus, the thinned cluster includes the initial points, that do not have neighboring points, and mean points representing subsets of the initial points. The result is shown in the third column of Fig.~\ref{fig:polyGeneration}. Compared to the second column containing the cluster of the polygonal chain, there are fewer points due to the reduction of data points.
\begin{center}
	\begin{algorithm}       
        \LinesNumbered        
		\SetFuncSty{textbf}  
        \SetKwInOut{KwParams}{Parameters}
		\SetKwFunction{getpoint}{getPointWithMostNeighbors} 
		\SetKwFunction{getneighbors}{getNeighbors} 
		\SetKwFunction{mean}{getMean}  	
  		\KwIn{pointsetIn}        
        \KwOut{pointsetOut}
        \KwParams{maxNeighborDist} 
        \SetKwProg{func}{Function}{}{}  
        \SetKw{break}{break}
        \SetKw{in}{Input}
        \SetKw{out}{Output}
        \SetKw{params}{Params}
  		\func{thinning(\in, \out, \params)}
  		{
        	pointsetOut = $\emptyset$\;
        	\While{true}
            {
              p = \getpoint(pointsetIn, maxNeighborDist)\;\label{line:getPoint}
              n = \getneighbors(p, maxNeighborDist)\;\label{line:neighbors}
              \If{n $\neq \emptyset$}
              {
              	\label{line:neighborAvailable}
              	mean = \mean(p $\cup$ n)\;\label{line:mean}
                pointsetIn = pointsetIn $\setminus$ \{p $\cup$ n\}\;\label{line:remove}
                pointsetOut = pointsetOut $\cup$ mean\;\label{line:add}
              }
              \Else
              {
              	\label{line:neighborNotAvailable}
              	\break\;
              }              			
            }
        	pointsetOut = pointsetOut $\cup$ pointsetIn\;\label{line:union}
  		}    
    \caption{Thinning step.}
    \label{algo:thinning}        
    \end{algorithm}
\end{center}

\paragraph{Polygon Generation}
Finally, the thinned point set is the input $pointsetIn$ for the last stage, in which the polygonal chain $polygon$ is generated (Algorithm~\ref{algo:polyGeneration}). This algorithm consists of two phases, i.e. forward and backward, and addresses the fifth challenge of the virtual border extraction step. Since a polygonal chain has a starting and ending point, we first select an arbitrary point of $pointsetIn$ as starting point and collect neighboring points in one direction. If there is no more neighboring point available, we again select our starting point and collect neighboring points in the other direction. Afterwards, the selected points are concatenated. In this context, direction corresponds to the sequence of the points of the polygonal chain. For example, considering an arbitrary point of a polygonal chain, which is not the starting or ending point, there are two directions from this point, i.e. the directions to the starting and ending point. To realize this behavior, we first initialize two empty polygonal chains $dir1$ and $dir2$ for each direction~(l.~\ref{line:poly:temp}), and we set the variable $forward$, that indicates the phase of the algorithm~(l.~\ref{line:poly:forward}). We then select an arbitrary point $p$ (here at index 0) of $pointsetIn$, mark it and append it to $dir1$~(l.~\ref{line:poly:start}ff.). Afterwards, the nearest neighboring point $n$ within a distance $maxNeighborDist$, that it not already marked, is selected~(l.~\ref{line:poly:neighbor}). If there is a neighboring point $n$ available~(l.~\ref{line:poly:notEmpty}), we append $n$ to one of the temporary polygonal chains depending on the variable $forward$, mark $n$ and select $n$ as the current point $p$~(l.~\ref{line:poly:append}ff.). This procedure is repeated until there is no neighboring point $n$ available for the current point $p$~(l.~\ref{line:poly:empty}), i.e. the neighboring points for the first direction are collected. In this case, we select our initial point again as current point $p$ and switch the variable $forward$ to collect neighboring points along the other direction~(l.~\ref{line:poly:change}ff.). Subsequently, the same procedure is performed until there is again no neighboring point $n$ available~(l.~\ref{line:poly:end}) or all points of $pointsetIn$ are marked~(l.~\ref{line:poly:loop}). As a last step, the order of the temporary polygonal chain $dir1$ is reversed~(l.~\ref{line:poly:reverse}), and $dir2$ is appended resulting in the final polygonal chain~(l.~\ref{line:poly:concat}). This is necessary to create a single polygonal chain with a single direction. The result is visualized in the last column of Fig.~\ref{fig:polyGeneration}.
\begin{center}
	\begin{algorithm}       
        \LinesNumbered        
		\SetFuncSty{textbf} 
        \SetKwInOut{KwParams}{Parameters} 
		\SetKwFunction{allMarked}{allMarked} 
		\SetKwFunction{getneighbor}{getNearestUnmarkedNeighbor} 
		\SetKwFunction{mark}{setMarked}  
        \SetKwFunction{reverse}{reverse} 
        \SetKwFunction{concat}{concat} 
  		\KwIn{pointsetIn}        
        \KwOut{polygon}
        \KwParams{maxNeighborDist}
        \SetKwProg{func}{Function}{}{}  
        \SetKw{break}{break}
        \SetKw{in}{Input}
        \SetKw{out}{Output}
        \SetKw{params}{Params}
  		\func{generatePoly(\in, \out, \params)}
  		{
        	dir1, dir2 = $\emptyset$\; \label{line:poly:temp}           
            forward = true\; \label{line:poly:forward}
            p = pointsetIn(0)\; \label{line:poly:start}
            \mark(p)\; \label{line:poly:mark1}
            dir1 = \concat(dir1, p)\; \label{line:poly:addP}
        	\While{not \allMarked(pointsetIn)}
            {        
            	\label{line:poly:loop}
              n = \getneighbor(p, maxNeighborDist)\; \label{line:poly:neighbor}
              \If{n $\neq \emptyset$}
              {
              \label{line:poly:notEmpty}
              	\If{forward}
                {
                	\label{line:poly:append}
                	dir1 = \concat(dir1, n)\;
                }
                \Else
                {
                	dir2 = \concat(dir2, n)\;
                } 
                \mark(n)\;
                p = n\;
              }
              \Else
              {
              \label{line:poly:empty}
              	\If{forward}
                {      
                	\label{line:poly:change}
                	p = pointsetIn(0)\;
                    forward = false\;
                }
                \Else
                {
                	\break;\label{line:poly:end}
                } 
              }              			
            } 
            \reverse(dir1)\;\label{line:poly:reverse}
            polygon = \concat(dir1, dir2)\;\label{line:poly:concat}
  		}    
    \caption{Polygon generation step.}
    \label{algo:polyGeneration}        
    \end{algorithm}
\end{center}
We use the parameter values shown in Tab.~\ref{tab:parameters} for the algorithms. These were determined experimentally.
\begin{table}[htbp]
  \centering
  \caption{Parameter values for the algorithms.}
    \begin{tabular}{l|lr}
    \toprule
    Stage  & Parameter & \multicolumn{1}{r}{Value}\\
    \midrule
    Clustering & eps   & 0.5 m \\
          & minPts & 1    \\
          & minExp & 0.3 m  \\
          & maxExp &    +$\infty$ m  \\
          & minSize & 10   \\
    \midrule
    Thinning & maxNeighborDist & 0.1 m  \\
    \midrule
    Polygon generation & maxNeighborDist & 0.5 m \\
    \bottomrule
    \end{tabular}%
  \label{tab:parameters}%
\end{table}%

\section{Experimental Evaluation}
In order to evaluate our proposed interaction method and our hypotheses concerning user requirements, such as interaction time, user experience, completeness and accuracy, we conducted an experimental evaluation involving multiple participants and three scenarios for restriction areas in a smart home environment. The evaluation was inspired by the USUS framework, that provides a methodological framework to evaluate certain aspects of a system involving the interaction of human and robot~\cite{Weiss:2009}.

\subsection{Independent Variables}
In our experiment, we manipulated a single independent variable, i.e. the interaction method. This variable can have one of the two values:
\begin{enumerate}
	\item \textbf{Robot only}: This is the baseline approach similar to Sprute et al.~\cite{Sprute:2019a}, that is based on sole HRI. A user interacts with the mobile robot by sketching the desired virtual border on the ground plane using a laser pointer. The mobile robot detects the laser spot employing its on-board camera and follows the laser spot using visual servoing technique if it attempts to leave the camera's field of view~\cite{Chaumette:2016}. In order to switch between different states of the interaction method, the user can push buttons on the backside of the robot or employ visual codes generated by the laser pointer. Feedback about the interaction method's current state is provided through on-board colored LEDs and sound when switching between states. Additionally, beep tones are uttered when the system acquires virtual border points~$\mathcal{P}$ or a seed point~$\bm{s}$.
    \item \textbf{Network robot system (NRS)}: This is our proposed approach based on a NRS described in Sect.~\ref{sec:approach}. Additional to the mobile robot, the NRS features stationary cameras in the environment as additional sensors to perceive laser points. A voice control allows switching between system's states using voice commands. Among colored LEDs and non-speech audio sound on board the mobile robot, a smart display integrated into the environment acts as additional feedback device. This shows a 2D OGM of the environment and the user-defined virtual borders.
\end{enumerate}

\subsection{Hypotheses}
The objective of the experimental evaluation was the test of the following hypotheses:
\begin{itemize}
    \item \textbf{Hypothesis 1}: Due to the extended perceptual capabilities, the interaction time is shorter when employing the \textit{NRS} compared to the \textit{Robot only} interaction method. We define the interaction time as the duration between the start, i.e. the first time employing the laser pointer, and the end of user interaction, i.e. the integration of the virtual border into the given map of the environment.
    \item \textbf{Hypothesis 2}: Due to the extended feedback and interaction capabilities, the user experience is better when employing the \textit{NRS} compared to the \textit{Robot only} interaction method. We define the user experience as a set of aspects concerning the interaction method, such as positive feeling, intuitiveness, demand, learnability and feedback. 
    \item \textbf{Hypothesis 3}: The completeness does not significantly get worse when employing the \textit{NRS} compared to the \textit{Robot only} interaction method. We define the completeness as the success rate at which a user successfully completes an interaction process. We register a successful interaction process if a participant can correctly specify the virtual border points $\mathcal{P}$ (independent of their accuracy) and can correctly specify a seed point $\bm{s}$ indicating the restriction area.
    \item \textbf{Hypothesis 4}: The accuracy does not significantly get worse when employing the \textit{NRS} compared to the \textit{Robot only} interaction method. We define the accuracy as the overlap between the user-defined virtual border as the result of the interaction process and the user-intended virtual border as required by the experimental scenarios.
\end{itemize}

\subsection{Setup}
\label{sec:setup}
\begin{figure*}
		\centering
        \begin{subfigure}[b]{0.48\textwidth}
                \centering
                \includegraphics[width=\textwidth]{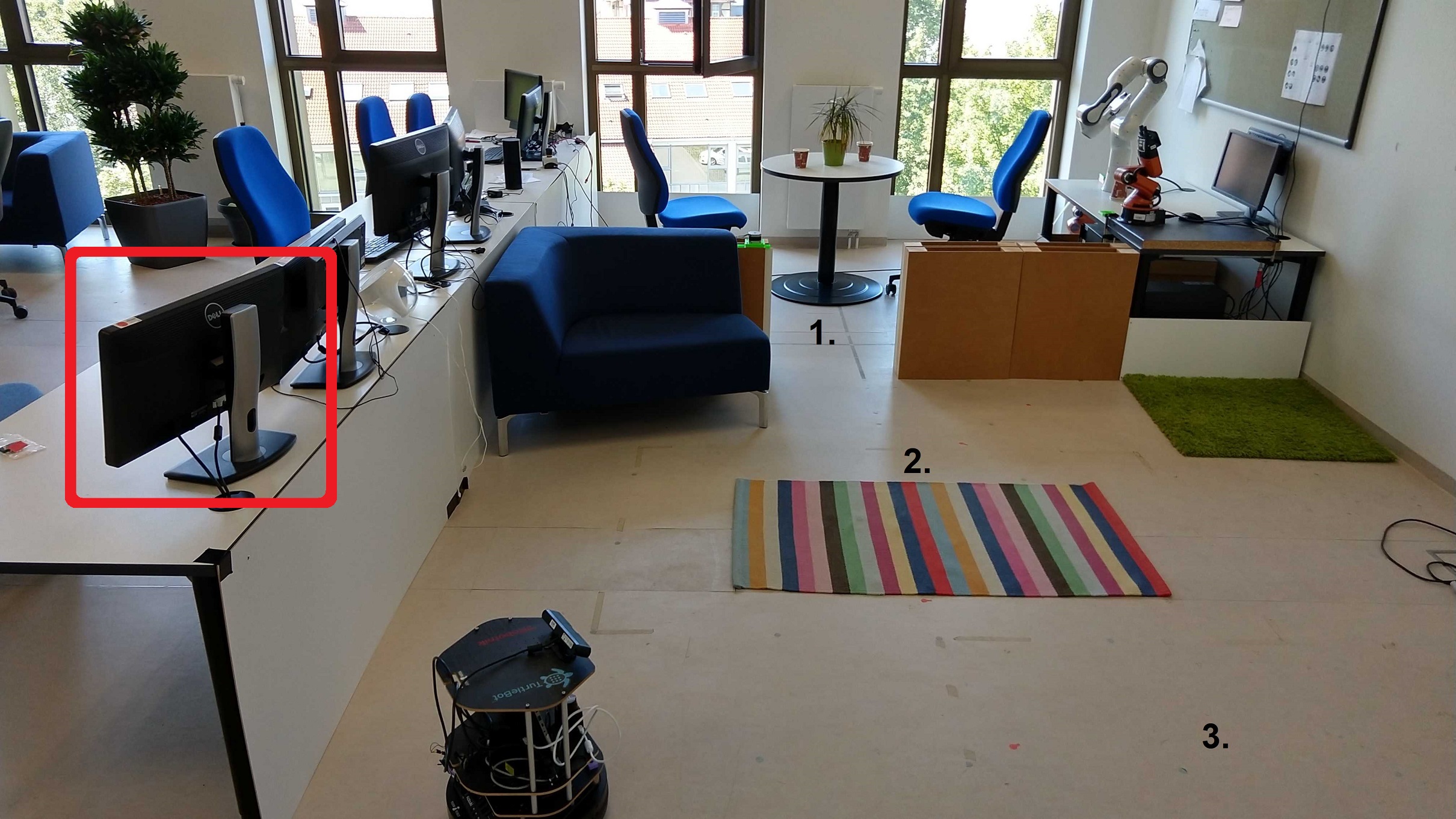}
                \caption{}    
                 \label{fig:labImage}                        
        \end{subfigure}        
        \centering
        \begin{subfigure}[b]{0.48\textwidth}
                \centering
                \includegraphics[width=\textwidth]{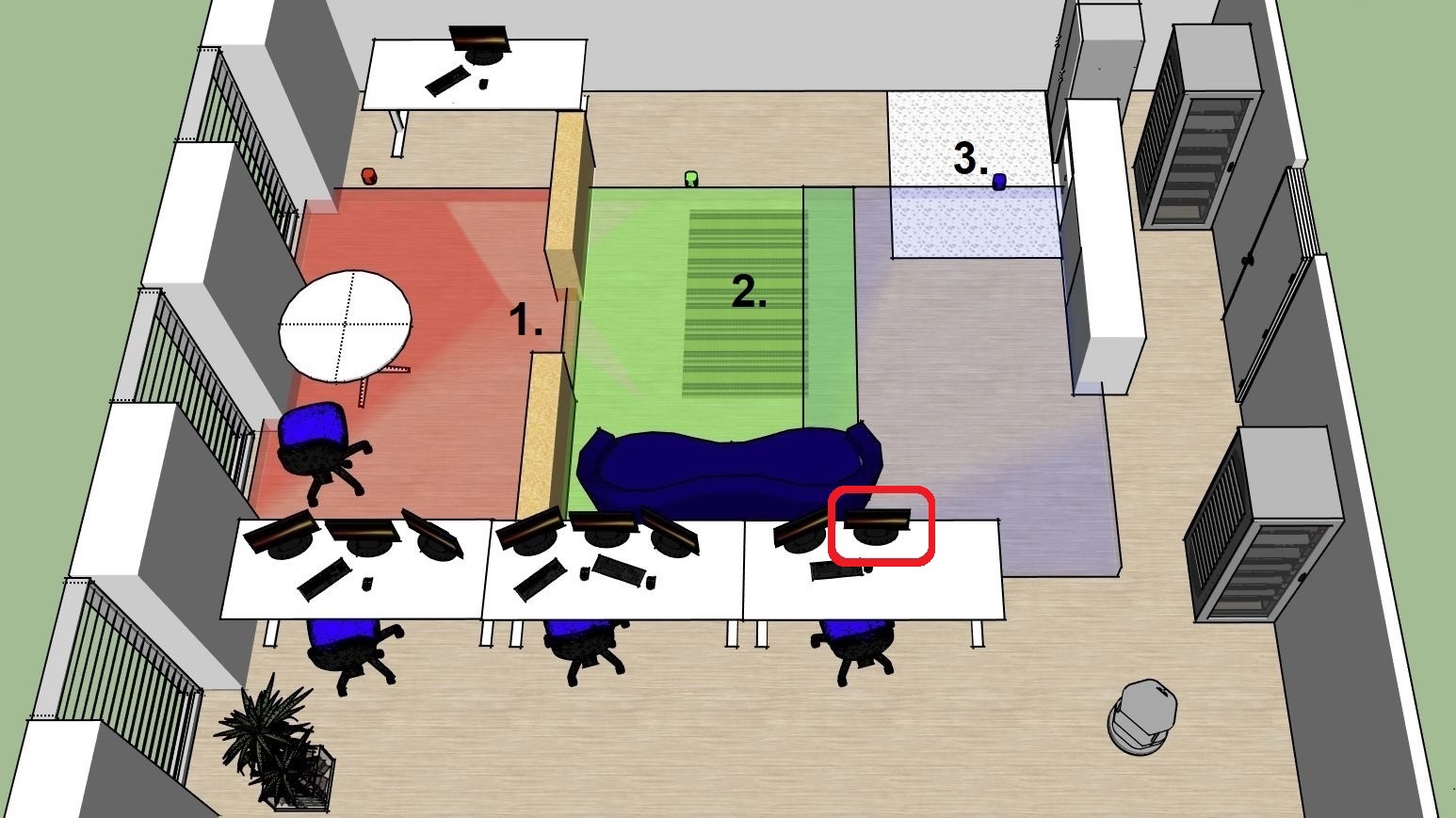}
                \caption{}       
                \label{fig:labSketch} 
        \end{subfigure}                  
        \caption{Image and 3D sketch of the lab environment. The three evaluation scenarios are numbered, and the three cameras' fields of view are visualized in red, green and blue. The position of the smart display for feedback is encircled in red, and the mobile robot's initial pose is depicted in the bottom right of the sketch.} 
        \label{fig:experimentalSetup}
\end{figure*}
To test our hypotheses, we motivated our experimental setup by three scenarios for restriction areas as presented in the introduction, i.e. (1)~privacy zones, (2)~carpets and (3)~dirty areas. For this purpose, we set up a home environment in our 8~m $\times$ 5~m lab environment comprising free space, walls, plants and furniture, such as tables and chairs. 
Additionally, we integrated adjustable walls with a height of 0.5~m to model two different rooms in the environment. Besides, we placed a carpet on the ground and some dirt in one area of the environment as basis for the evaluation scenarios. An image and 3D sketch of the environment is shown in Fig.~\ref{fig:experimentalSetup}.

As a mobile robotic platform, we employed a TurtleBot v2 equipped with a laser scanner for localization and a front-mounted RGB-D camera. This is a typical evaluation platform whose mobile basis is similar to a vacuum cleaning robot with differential-drive wheels. The camera's color images were captured with a resolution of 640~$\times$~480~pixels, and the depth sensor had a resolution of 160 $\times$ 120~pixels. Additionally, the robot had a colored on-board LED, three push buttons and a speaker. A prior OGM of the environment (2.5~cm per cell) was created beforehand using a Simultaneous Localization and Mapping (SLAM) algorithm running on the mobile robot~\cite{Grisetti:2007}, and the mobile robot was localized inside the environment using the adaptive Monte Carlo localization approach~\cite{Fox:2003}. This allows the robot to determine its 3-DoF pose, i.e. position and orientation, with respect to the map coordinate frame.

In order to take advantage of a smart home environment as described in Subsect.~\ref{sec:desgin}, we mounted three RGB cameras with an image resolution of $1920 \times 1080$ pixels on the ceiling (2.95~m height, pitch angle of 90$^\circ$). Thus, they provided top views of the environment. Their fields of view partly overlapped as illustrated in Fig.~\ref{fig:labSketch}, but they did not cover the entire environment. Hence, there was only a partial observation of the environment, which is typical for home environments.  All RGB cameras were calibrated, i.e. their intrinsic camera parameters were known, and their relative transformations with respect to the map coordinate frame were determined in advance. We denote these stationary cameras as \textit{red, green and blue} camera. The interaction using speech commands relied on a Wizard of Oz method in which a human operator reacted on the speech commands of participants, i.e. switching between system's states per remote control. We did not use a voice-controlled intelligent personal assistant, such as Amazon's Alexa or Google Assistant, due to network restrictions in the university's network. However, this method was not recognized by the participants and did not change the way in which a participant interacted with the system. Moreover, we placed a 22-inch smart display on a table near the restriction areas to provide visual feedback to the participants. This display was network-connected to the system and showed the progress of the interaction process, i.e. the OGM of the environment, the mobile robot's current pose and virtual borders if specified by the user. 

\subsection{Procedure}
\label{sec:procedure}
After setting up the experimental environment, each participant was introduced to the following three evaluation scenarios. These scenarios covered both types of virtual borders and are good representatives for restriction areas:
\begin{enumerate}
	 \item \textbf{Room exclusion}: The user wants the mobile robot to not enter a certain room due to privacy concerns. For this purpose, the user draws a line separating the room from the rest of the environment. This area is in the fields of view of the red and green camera. The length of the polygonal chain $\mathcal{P}$ is 0.70~m long, and the area has a size of approximately 8.00~m$^2$. 
	\item \textbf{Carpet exclusion}: The user wants the mobile robot to circumvent a carpet area (2.00~m $\times$ 1.25~m) while working. To this end, he or she draws a polygon around the carpet and specifies the inner area as restriction area. This area is in the fields of view of the green and blue camera.
    \item \textbf{Spot cleaning}: The user wants his or her vacuum cleaning robot to perform a spot cleaning in a corner of a room. Hence, he or she draws a separating curve around the area and specifies the rest of the room as restriction area. This dirty area is indicated by paper snippets on the ground and is partly covered by the blue camera. The polygonal chain has a length of 3.60~m and encompasses an area of 3.20~m$^2$.
\end{enumerate}
In addition to the camera view coverage of the restriction areas as described, it was possible that participants temporarily occluded restriction areas with their bodies depending on their positions during interaction.

In this experiment, we applied a within-subjects design, i.e. each participant evaluated both interaction methods in all three evaluation scenarios. The order of the selected interaction method was randomized to avoid order effects. After selecting an interaction method, an experimenter explained a participant how to employ the selected interaction method, i.e. a short introduction into the different states of the system, how to switch between states and how to use the laser pointer. Afterwards, a participant had some time to get familiar with the interaction device, i.e. the participant could use the laser pointer and guide the mobile robot. This took approximately five minutes. Subsequently, the participant started to specify virtual borders for the three scenarios. The order of the different scenarios was randomized. At the beginning of each scenario, the mobile robot's initial pose was set to a predefined pose to allow the comparison between the results. This initial pose, that was not covered by a stationary camera, is shown in the bottom right of Fig.\ref{fig:labSketch}. 

The shortest paths to the restriction areas in the three scenarios were between 2.50~m and 5.40~m (Scenario~1: 5.40~m, Scenario~2: 2.50~m and Scenario~3: 3.00~m). Since we applied a within-subjects design, this procedure resulted in six runs per participant (three scenarios and two interaction methods) and a participant could compare both interaction methods. After performing the practical part of the experiment, each participant was asked to answer a post-study questionnaire concerning his or her user experience with the interaction methods. An experiment with a single participant took approximately 20 minutes in total.

\subsection{Measurement Instruments}
\label{sec:instruments}
During the experiment, the time needed to specify a virtual border for each participant, interaction method and scenario was measured. Thus, we obtained six measurements for each participant. The time measurement started with the change from the \textit{Default} state to one of the active states and ended with the integration of the virtual border into the prior map. Moreover, the time measurement was decomposed according to the three active states of the system, i.e. guiding the mobile robot ($Guide$), specifying virtual border points ($Border$) and the seed point ($Seed$), to identify reasons for potential time differences between the interaction methods.

In addition to the time, an experimenter documented if a participant successfully specified a virtual border for an evaluation scenario, i.e. a participant could specify the virtual border points $\mathcal{P}$ (independent of their accuracy) and a seed point $\bm{s}$. The success values are used to assess the completeness of an interaction method. To this end, we calculate the ratio between the number of successful runs and the total number of runs for each interaction method and scenario.

To measure the user experience, participants were asked to answer a post-study questionnaire. In addition to general information, such as age, gender and experience with robots, the questionnaire contained five statements concerning the user experience, that could be rated on 5-point Likert items with numerical response format. This questionnaire was inspired by the questionnaire of Rouanet et al., who used a similar questionnaire to assess the usability and user experience of human-robot interfaces~\cite{Rouanet:2013}. The questionnaire included the following statements about the experiment (translated from German): 
\begin{enumerate}
	\item I had problems to define the virtual borders\\ (1 = big problems, 5 = no problems)
    \item It was intuitive to define the virtual borders\\ (1 = not intuitive, 5 = intuitive)
    \item It was physically or mentally demanding to define the virtual borders (1 = hard, 5 = easy)
	\item It was easy to learn the handling of the interaction method (1 = hard, 5 = easy)
	\item I liked the feedback of the system\\ (1 = bad/no feedback, 5 = good feedback)
\end{enumerate}
Furthermore, the participants were asked if the smart environment supported the interaction process. They could answer the question with \textit{yes} or \textit{no}.

The fourth evaluation criterion is the accuracy, that we assessed by calculating the Jaccard similarity index (JSI) between a user-defined $UD$ and its associated ground truth $GT$ virtual border:
\begin{equation}
JSI(GT, UD) = \dfrac{|GT \cap UD|}{|GT \cup UD|} \in [0, 1]
\end{equation}
This value indicates the overlap between two virtual borders. To calculate this overlap, a ground truth map $GT$ was manually created for each scenario in advance, which contained the physical environment as well as the virtual border. After each run of the experiment, the resulting user-defined map $UD$ was automatically saved and associated with its ground truth map $GT$ for evaluation of the accuracy. 

\subsection{Participants}
The experiment was conducted with a total of 15 participants (11 male, 4 female) with a mean age of $M=28.8$ years and standard deviation of $SD=11.4$ years. The age group ranged from 17 to 55 years. Participants were recruited from the local environment by word of mouth and rated their experience with robots on a 5-point Likert item ranging from \textit{no experience}~(1) to \textit{highly experienced}~(5) with a mean of $M=3.2$ and standard deviation of $SD=1.4$. This corresponds to a moderate experience with robots and comprises users owning a mobile robot, such as a vacuum cleaning robot. However, they only deploy the mobile robots in their home environments according to the manual and do not know how they internally work.

\subsection{Tools}
We implemented all components of the system as ROS nodes~\cite{Quigley:2009}. ROS is a modular middleware architecture that allows communication between several components of a system, that are called nodes. ROS is the de facto standard for robot applications and provides a large set of tools to accelerate prototyping and error diagnosis. We organized all our nodes in packages to allow the easy distribution of our implementation. In order to save resulting maps of an interaction process for evaluation purposes, we implemented a node that directly stores the map on the hard disk whenever a new map is defined by a participant. Furthermore, we used integrated time functionality of ROS to perform time measurements for the evaluation. The visualization on the smart display was based on RVIZ, which is a 3D visualization tool for ROS.

\subsection{Analysis \& Results}
To test our hypotheses, we report the analysis and results of the experimental evaluation in this subsection. In case of statistical tests to identify differences between the interaction methods, we chose a significance level of $\alpha=.05$.

\subsubsection{Interaction Time}
\label{sec:teachingTime}
\begin{figure}
	\centering
	\includegraphics[width=\columnwidth]{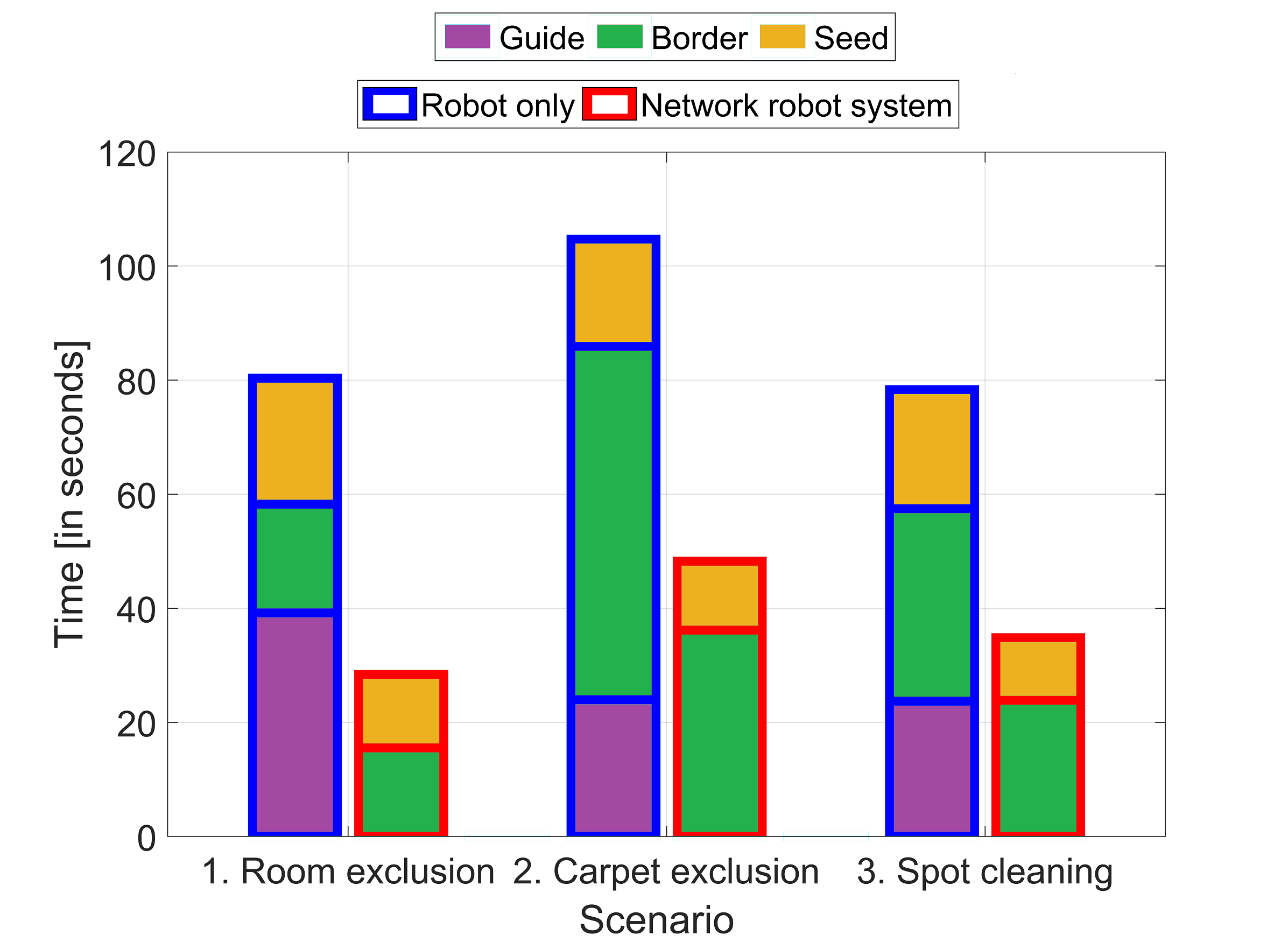}   
	\caption{Interaction time for both interaction methods (blue and red contours around bars) based on the scenarios. Each bar is subdivided into the three active states of the system.}
	\label{fig:time}
\end{figure}
The interaction times of the experiment are summarized in Fig.~\ref{fig:time}. Each bar comprises the measurements of all 15 participants for an interaction method and scenario. Hypothesis~1 stated that the interaction time would be reduced when using the \textit{NRS} compared to the \textit{Robot only} interaction method. To test this hypothesis, we first ran Shapiro-Wilk tests to test for normality of the data (differences in the interaction times between the two interaction methods), which is an assumption of parametric statistical hypothesis tests, e.g. a paired \mbox{t-test}. These tests only became significant for the third scenario ($p=.016$). Thus, we assume the data of the first two scenarios to be approximately normally distributed, while the third scenario is not normally distributed. Moreover, we interpreted the corresponding boxplots for outliers ($1.5 \times interquartile\ range$). The boxplots revealed that there were some outliers in the data. Due to the presence of outliers and the violation of normality in the third scenario, we performed a non-parametric Wilcoxon signed-rank test to compare both interaction methods. The statistical results show that there is a significant difference between the interaction methods in all evaluation scenarios:
\begin{itemize}
	\item 1. Room exclusion:\tab $Z = -3.411,\ p < .001$
    \item 2. Carpet exclusion:\tab $Z = -3.408,\ p < .001$
    \item 3. Spot cleaning:\tab $Z = -3.409,\ p < .001$
\end{itemize}
Our proposed \textit{NRS} approach is significantly faster compared to the baseline approach. This results in speedups of 2.8, 2.2 and 2.2 for Scenarios 1, 2 and 3. 

The reason for this significant difference is revealed by the decomposition of the time measurements. While the time for the \textit{Robot only} approach is composed of all active states of the system, our \textit{NRS} approach does not include the \textit{Guide} state. This is a consequence of the human-robot-environment interaction where the \textit{NRS} automatically sends the mobile robot to the intended restriction area when a laser spot is detected by a stationary camera. Therefore, the user is not restricted to the mobile robot's line of sight and does not have to manually guide the mobile robot to the intended restriction area. Thus, the \textit{NRS} approach can avoid the time in the \textit{Guide} state. This time primarily depends on the distance between the robot's start pose and the restriction area. As described in Subsect.~\ref{sec:procedure}, the distances in our scenarios ranged from 2.5~m to 5.4~m. If the distances would be smaller, the time in the \textit{Guide} state would also decrease. However, we think that we chose the distances quite liberally since much larger distances would be even realistic, e.g. in typical home environments with a single charging station for the mobile robot. 

Another reason for the time difference between the interaction methods is the time in the state \textit{Border}, which is linear with respect to the border length~\cite{Sprute:2019a}. Thus, if the user-defined virtual border is short, e.g. 0.70~m for Scenario~1, our \textit{NRS} approach is only slightly faster in this state, i.e. 4 seconds difference. But if we consider a larger virtual border, e.g. the 6.50~m long border around the carpet (Scenario~2), our approach is even 26 seconds faster on average. The reason for this is the mobile robot's velocity limitation (0.2~$m/s$) to ensure a safe and smooth movement of the robot. By using our \textit{NRS} approach, this speed limitation can be reduced if the laser spot is in the field of view of one of the stationary cameras. Our interaction method is then only limited by the detection rate of the cameras (25~$frames/s$). Hence, it also features a linear interaction time, but with a smaller gradient. 

Another speedup is achieved when specifying the seed point (\textit{Seed} state) because a user can directly indicate the seed point with laser points, which are detected by a stationary camera. In case of the \textit{Robot only} approach, a user additionally  has to rotate the mobile robot around its vertical axis to adjust the camera's field of view. This rotation takes additional interaction time. In summary, the results support Hypothesis~1.

\subsubsection{User Experience}
\begin{figure}
	\centering
	\includegraphics[width=\columnwidth]{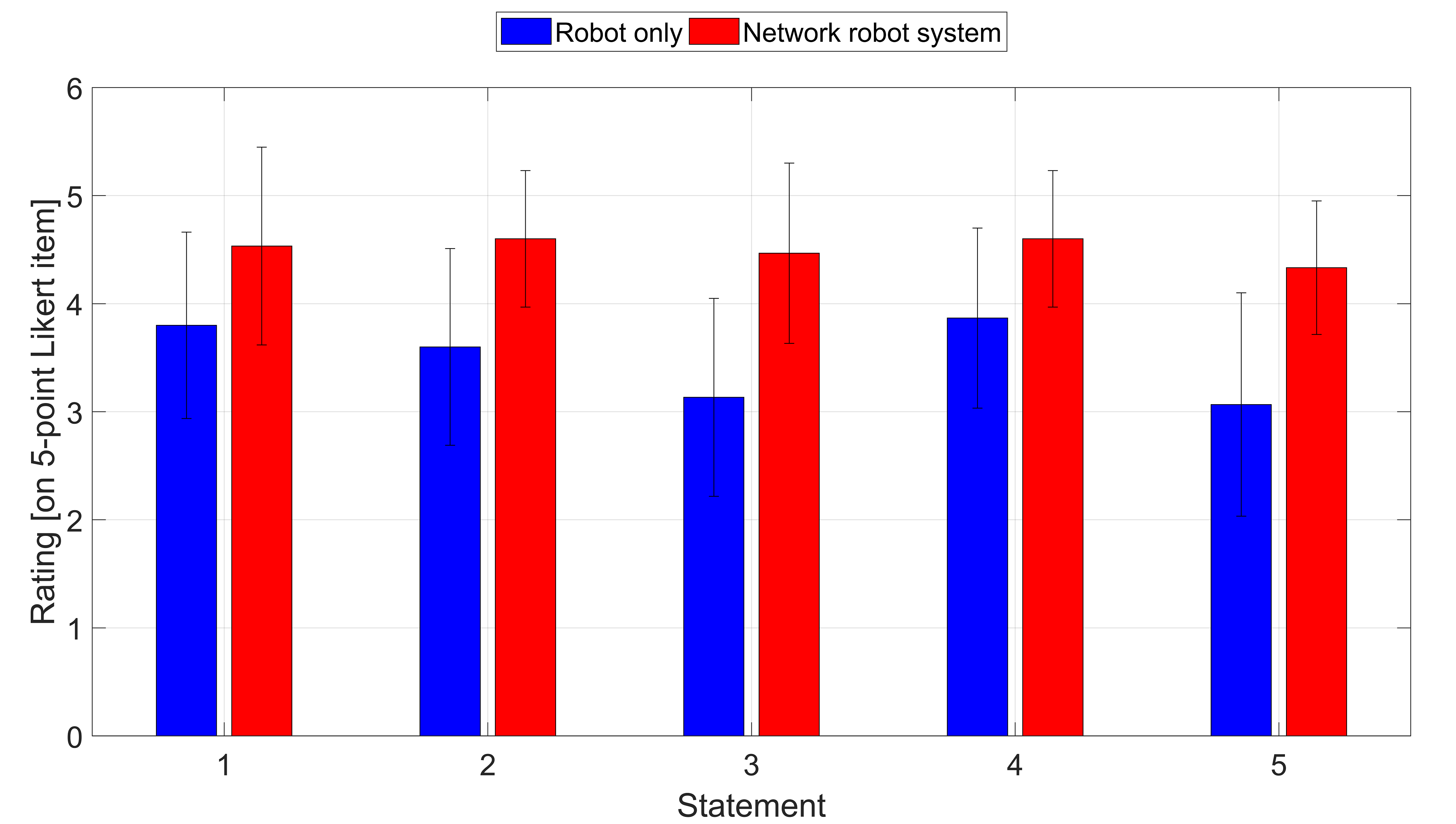}   
	\caption{User experience for both interaction methods showing mean and standard deviation depending on the statements.}
	\label{fig:userExperience}
\end{figure}

In order to assess the user experience and test Hypothesis~2, i.e. the user experience is better when employing the \textit{NRS} compared to the \textit{Robot only} interaction method, we considered the participants' answers of the questionnaire introduced in Subsect.~\ref{sec:instruments}. The participants' answers to the questionnaire are visualized in Fig.~\ref{fig:userExperience} with their mean ($M$) and standard deviation ($SD$) per statement and interaction method.

\begin{figure}
	\centering
	\includegraphics[width=\columnwidth]{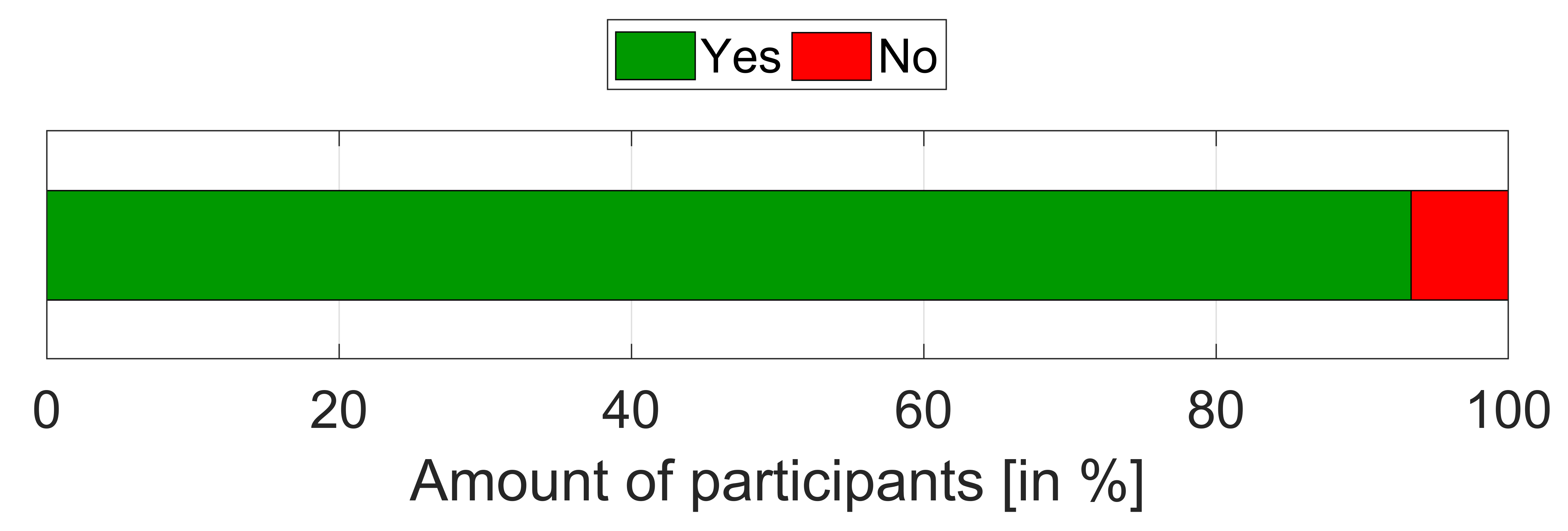}   
	\caption{Answers of the participants to the question if the smart environment supported the interaction process.}
	\label{fig:help}
\end{figure}

Since Likert-item data violate the assumption of normality, we ran a Wilcoxon signed-rank test on the 15 answers per statement to determine whether there are statistically significant differences between both interaction methods. We found statistically significant differences for all statements in the questionnaire:
\begin{itemize}
	\item Statement 1:\tab $Z = -3.051,\ p$ = $.002$
    \item Statement 2:\tab $Z = -2.830,\ p$ = $.004$
	\item Statement 3:\tab $Z = -3.126,\ p$ < $.001$
	\item Statement 4:\tab $Z = -2.373,\ p$ = $.023$
	\item Statement 5:\tab $Z = -2.745,\ p$ = $.005$
\end{itemize}
Our proposed interaction method based on a \textit{NRS} is better rated by the participants for all statements compared to the baseline approach. Participants had significantly less problems defining the virtual borders with a \textit{NRS} ($M=4.53$) than without support of a smart environment ($M=3.80$)\footnote{Although the median should be used to describe the central tendency in non-parametric tests, we report the mean value in this paragraph to better reveal the differences between the interaction methods. This is valid since we consider an interval-level of measurement~\cite{Harpe:2015}. Moreover, since other studies often report mean values for Likert-items, we also want to make our results better comparable.}. A reason could be that some participants had problems to rotate the mobile robot around its vertical axis, e.g. to specify the seed point. In this case, they moved the laser spot too fast so that the robot's on-board camera could not follow the spot on the ground. In contrast to this, our approach avoids this problem by the additional cameras in the environment. Participants also found our approach more intuitive ($M=4.60$ and $M=3.60$) because they could more directly specify a virtual border without concentrating on the mobile robot. This was automatically sent to the restriction area when a stationary camera detected a laser spot. Furthermore, the speech commands provided a more intuitive interaction medium to change the system's internal state than pushing on the robot's buttons. The strongest effect was measured for Statement~3 that shows that our approach is less physically or mentally demanding ($M=4.47$) compared to the baseline approach ($M=3.13$). This coincides with the results of the interaction time as reported in Subsect.~\ref{sec:teachingTime}. Another difference was observed for the learnability of the interaction methods (\textit{NRS}: $M=4.60$ and \textit{Robot only}: $M=3.87$). Since both interaction methods are based on a laser pointer as interaction device, the handling of a laser pointer does not influence the learnability. However, it could be easier to learn the speech commands than the assignments of the buttons to change between states of the system. Furthermore, the guiding of the mobile robot could affect the rating. Finally, there is a significant difference for the feedback of the interaction method ($M=4.33$ and $M=3.07$). Since the feedback system of the NRS is a superset of the baseline's feedback system, a major reason for this difference is the additional smart display that visualizes the OGM with the user-defined virtual borders. This kind of feedback is missing for the \textit{Robot only} approach that only features sound and colored LED feedback. The answers to the question if the smart environment supported the interaction process are summarized in Fig.~\ref{fig:help}. Most of the participants (14 out of 15) felt that the smart environment supported the interaction process. In summary, Hypothesis~2 is supported by the results.

\subsubsection{Completeness}
\begin{figure}
	\centering
	\includegraphics[width=\columnwidth]{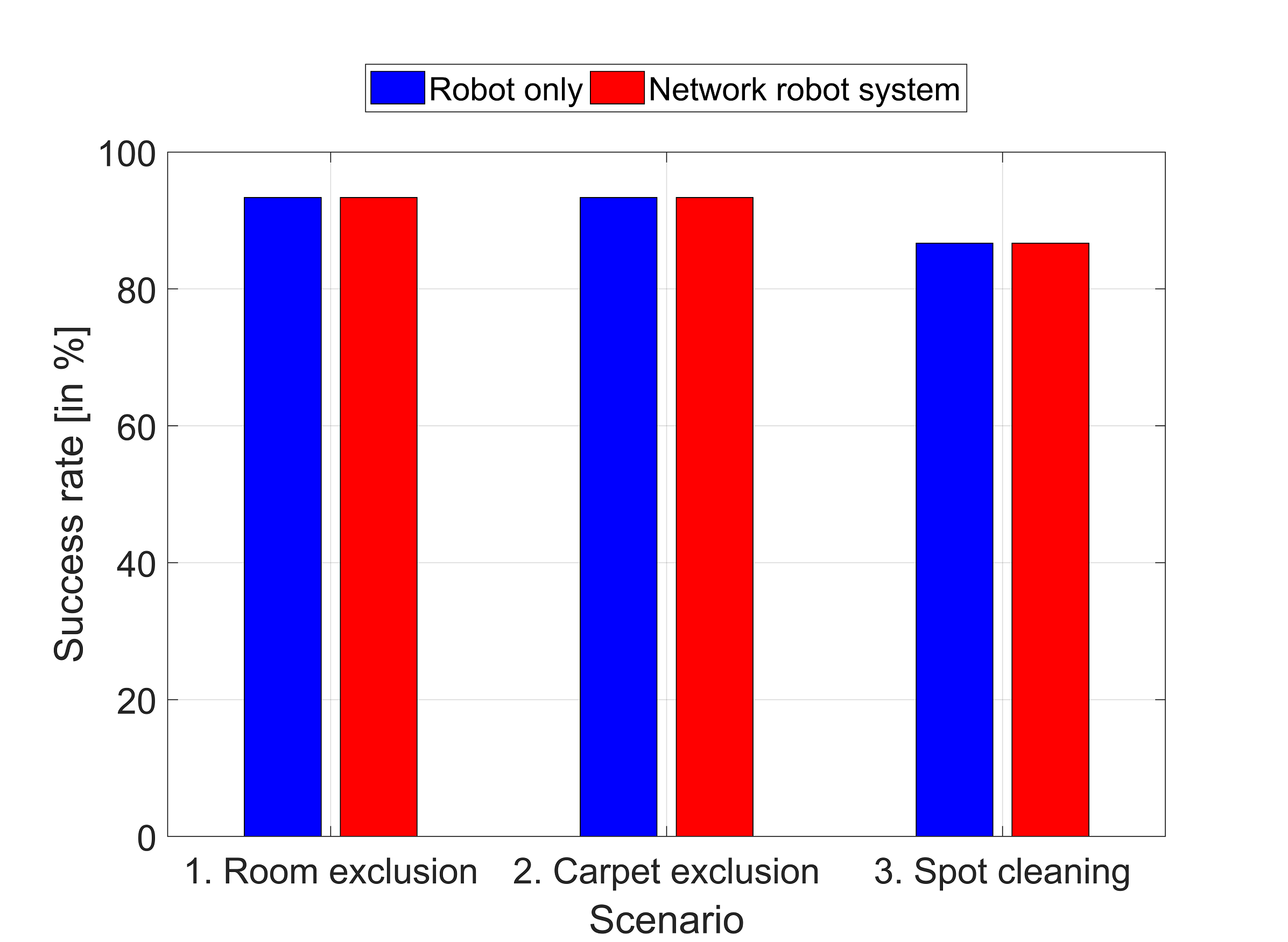}   
	\caption{Completeness for both interaction methods depending on the scenarios.}
	\label{fig:success}
\end{figure}

Hypothesis~3 stated that the completeness does not significantly get worse when employing the \textit{NRS} compared to the \textit{Robot only} interaction method. The success rates are summarized in Fig.~\ref{fig:success} as bars. Both interaction methods feature the same high success rate, i.e. 91.1\% on average. Furthermore, the success rates for the three scenarios are the same (93.3\% for Scenarios~1 and 2; 86.7\% for Scenario~3). There were nine participants who performed all their runs successfully, four participants who failed for one of their six runs, and two participants incorrectly defined a virtual border in two of their six runs. The reason for these incorrect runs was the definition of the seed point $\bm{s}$. While some participants were confused where to specify the seed point~$\bm{s}$, especially in Scenario~3, other participants were unfocused and noticed their mistake on their own after performing the experiments. There were no problems with the definition of the virtual border points $\mathcal{P}$. Since both interaction methods feature an equally high completeness, we assume Hypothesis~3 to be supported by the results. 

\subsubsection{Accuracy}
\label{sec:evalAccuracy}
\begin{figure}
	\centering
	\includegraphics[width=\columnwidth]{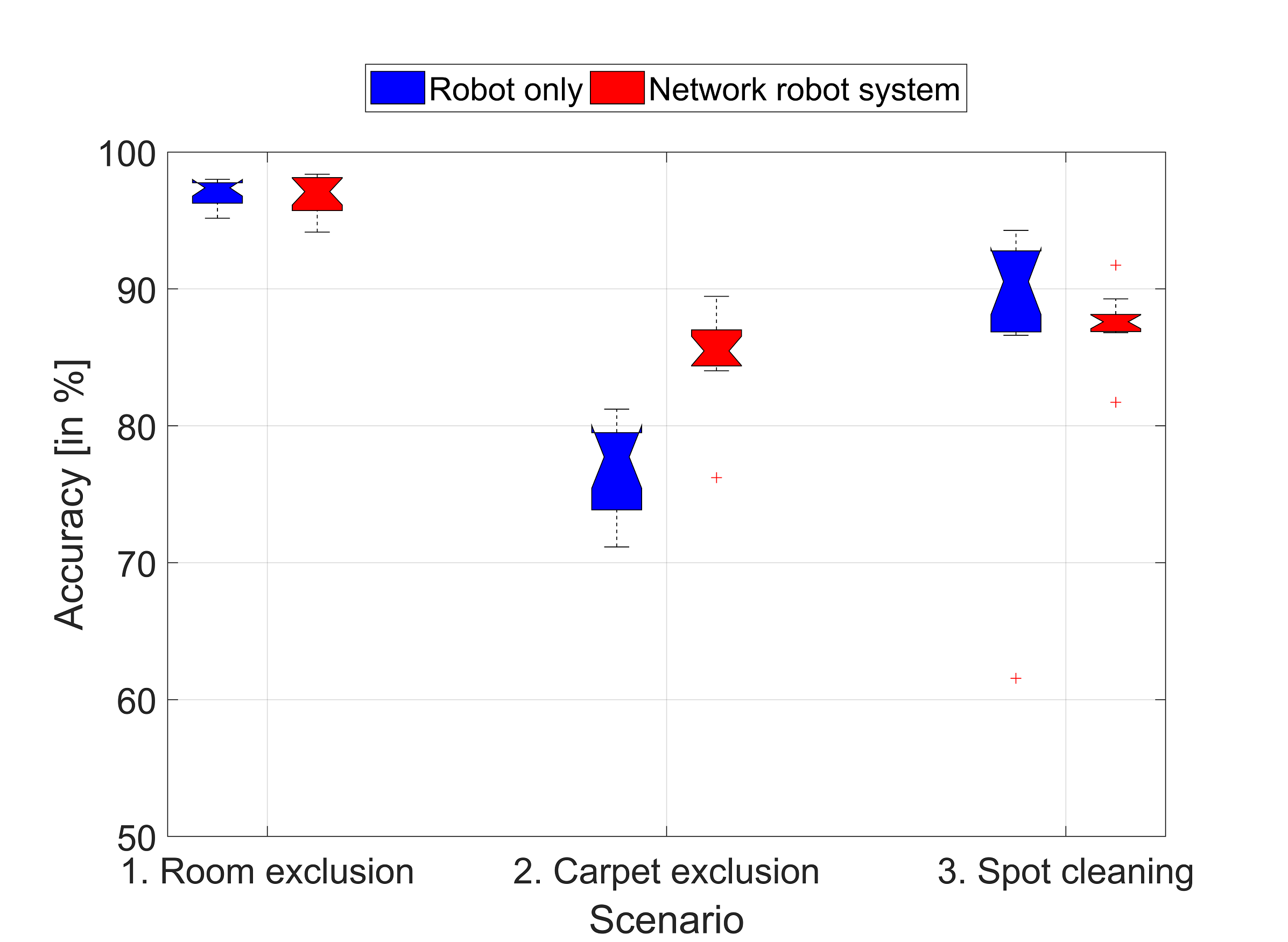}   
	\caption{Quantitative accuracy results for both interaction methods depending on the scenarios.}
	\label{fig:accuracy}
\end{figure}
Finally, Hypothesis~4 stated that the accuracy does not significantly get worse when employing the \textit{NRS} compared to the \textit{Robot only} interaction method. The JSI values resulting from the experimental evaluation are presented in Fig.~\ref{fig:accuracy}. To test the hypothesis, we again explored the data for normality and outliers using Shapiro-Wilk test and boxplot interpretation. The Shapiro-Wilk tests did not become significant for any scenario indicating a normal distribution of the data. However, since the data for Scenarios~2 and 3 contain some outliers, we preferred a Wilcoxon signed-rank test to a paired t-test. The statistical results were different for the three scenarios:
\begin{itemize}
	\item 1. Room exclusion:\tab $Z = -0.594,\ p$ = $.588$
    \item 2. Carpet exclusion:\tab $Z = -3.110,\ p$ < $.001$
    \item 3. Spot cleaning:\tab $Z = -2.497,\ p$ = $.010$
\end{itemize}
\begin{figure*}
	\centering
	\includegraphics[width=\textwidth]{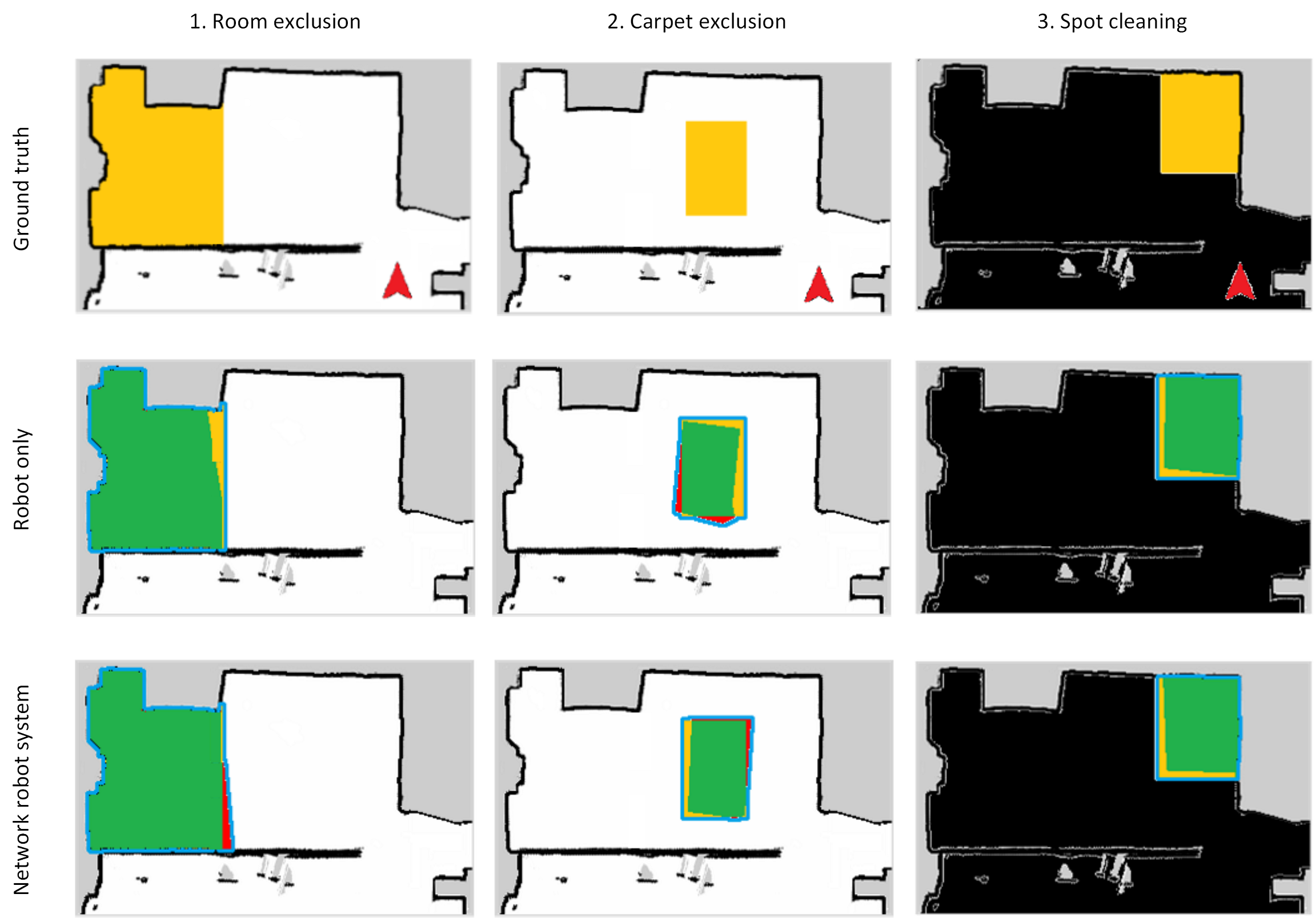}   
	\caption{Visualization of representative accuracy results. The first row shows ground truth maps for all three scenarios containing occupied (black), free (white) and unknown (gray) areas. The yellow cells indicate the area to be specified by a participant during an interaction process~(GT). The robot's start pose is visualized as a red arrow. The following rows visualize the qualitative accuracy results for both interaction methods. A user-defined area as result of an interaction process is colored in green and red. Green pixels indicate the overlap of ground truth and user-defined areas (GT $\cap$ UD), while red pixels show areas defined by the user but not contained in the ground truth map (UD $\setminus$ GT). A blue contour surrounds the union set of both areas (GT $\cup$ UD). Colors are only used for visualization.}
	\label{fig:accVisualization}
\end{figure*}
There is no significant difference for Scenario~1 (\textit{NRS}: $Mdn = .971$, \textit{Robot only}: $Mdn = .974$), the \textit{NRS} ($Mdn = .856$) performs significantly better than the \textit{Robot only} ($Mdn = .779$) in Scenario~2 and the \textit{Robot only} ($Mdn = .906$) has a significant higher accuracy than our \textit{NRS} ($Mdn = .878$) in Scenario~3. However, although the results are significant for Scenario~3, the difference between the medians is only $.028$, which is not notable in practice. This can also be qualitatively observed in Fig.~\ref{fig:accVisualization}, where the JSI can be visually interpreted as the fill degree of the green cells with respect to the cells encompassed by the blue contour. The results depicted in the figure are representative examples of different participants whose accuracy values coincide with the overall results per scenario and interaction method. The accuracy for Scenario~1 and 3 is similar, but there is a small difference for Scenario~2. The \textit{NRS} interaction method features a higher fill degree of the green area with respect to the area surrounded by the blue contour compared to the \textit{Robot only} approach. A reason could be that it is harder to specify the carpet's corners when guiding the robot compared to our \textit{NRS} approach. The highest accuracy value is achieved in Scenario~1 because the user excludes a relatively large area compared to the length of the virtual border. Thus, there is only minimal room for errors. In summary, the accuracy results support Hypothesis~4 because there is no significant degradation of the accuracy when employing the \textit{NRS} approach.

\subsection{Discussion}
We designed our experimental evaluation according to the idea that the environment is partially observed by stationary cameras and that there are typical restriction areas, such as carpets or privacy zones. This is a typical setting for a smart home environment. In this case, our results suggest that a smart environment can significantly improve the interaction time (Hypothesis~1) and user experience (Hypothesis~2) while not negatively affecting the completeness (Hypothesis~3) and accuracy (Hypothesis~4). 

However, there are two important aspects, that influence the speedup in the interaction time: (1)~the stationary cameras' coverage of the environment and (2)~the distance between the mobile robot's start pose and the restriction area. If we would decrease the number of cameras in the environment and thus the camera coverage, this would result in a smaller reduction of the interaction time. This would finally degenerate to the baseline approach if there is no camera coverage. Moreover, the speedup strongly depends on the distance between the mobile robot's start pose and the restriction area, which influences the performance of the baseline approach. This is due to the fact that the baseline approach requires a direct line of sight between human and robot, and thus a human first has to guide the mobile robot to the restriction area. For these reasons, it is not possible to report a specific speedup value. Nonetheless, in the evaluation scenarios we chose a typical camera coverage and reasonable distances, which are typical for home environments with a single charging station for a mobile robot. Hence, we conclude that the interaction time improves with the support of a smart environment, but we cannot report a specific speedup value. Therefore, the reported speedups in Subsection~\ref{sec:teachingTime} are intended to give an estimate and are only valid for this specific experimental evaluation.

In case of the user experience, participants appreciated aspects, such as a reduced mental or physical demand during interaction and an improved feedback of the system. These are effects of the incorporation of the smart environment's components. However, a participant also wished an even stronger feedback system. The participant did not like the change of attention between the sketching of the virtual border on the ground and the view on the smart display, which was positioned aside on a table. Therefore, we conclude that our choice of smart home components improved the user experience significantly but the incorporation of more expressive feedback channels could improve the user experience even more.

\section{Conclusions \& Future Work}
In this work, we investigated the effect of a smart home environment on the interaction process of restricting a mobile robot's workspace. For this purpose, we developed a novel interaction method leveraging a smart home environment and laser pointer in the interaction process. This NRS was designed to support the interaction process in terms of perception and interaction abilities. To this end, we first selected smart home components that we employed to realize a cooperative behavior between human, robot and smart environment in the interaction process. Especially, the cooperative perception involving stationary and mobile cameras to perceive laser spots and an algorithm to extract virtual borders from multiple camera observations is a major contribution of this work. This cooperation between mobile robot and smart environment neutralized the mobile robot's perceptual and interaction limitations. This was supported by an experimental evaluation that revealed an improvement of interaction time and user experience compared to a baseline method while preserving the high completeness and accuracy of the baseline. Hence, we conclude that a smart environment can improve the interaction process of restricting a mobile robot's workspace.

Nonetheless, the experimental evaluation also revealed current limitations of this work. For example, a participant wished a stronger feedback system without change of attention between the restriction area on the ground and the map shown on the smart display. For this purpose, an augmented reality solution would be promising, e.g. projectors integrated into the smart environment could visualize the user-defined virtual borders directly on the ground without change of attention. Although there are already solutions in the industrial context~\cite{Leutert:2013}\cite{Ganesan:2018}, projectors are not yet widely distributed in current smart homes. Moreover, we currently evaluated our interaction method with a single mobile robot, which is valid for most households. However, it would be interesting to incorporate multiple robots into the interaction process to increase the number of mobile cameras. Our proposed architecture already considers this aspect (Fig.~\ref{fig:overview}), but more work on the cooperation between multiple mobile robots in such a scenario is needed.

\section*{Acknowledgement}
This work is financially supported by the German Federal Ministry of Education and Research (BMBF, Funding number: 13FH006PX5). We would also like to thank all participants who took part in the experimental evaluation for their time and valuable feedback.



\bibliographystyle{ios1}           
\bibliography{bibo}        		

\end{document}